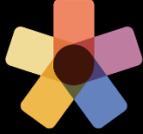

# Evo* 2023

## The Leading European Event on

## Bio-Inspired Computation

*Brno (Czech Republic). 12-14 April 2023*

# – LATE-BREAKING ABSTRACTS –

**Editors:**

**A.M. Mora**
**A.I. Esparcia-Alcázar**

# Preface

This volume comprises the Late-Breaking Abstracts accepted for the Evo* 2023 Conference, hosted in Brno (Czech Republic), from April 12th to 14th.

These abstracts were featured in both short talks and the conference's poster session, offering insights into ongoing research and preliminary findings exploring the application of various Evolutionary Computation approaches and other Nature-Inspired techniques to real-world problems.

These contributions represent promising developments, highlighting forthcoming advances and applications in the field of nature-inspired methods, particularly Evolutionary Algorithms.

*Antonio M. Mora*
*Anna I. Esparcia-Alcázar*

# Table of contents

# Multi-objective Design of Hardware Accelerators for Levodopa-Induced Dyskinesia Classifiers

Martin Hurta[(✉)][0000−0002−7915−8147], Vojtech Mrazek[0000−0002−9399−9313],
Michaela Drahosova[0000−0003−1217−4609], and
Lukas Sekanina[0000−0002−2693−9011]

Faculty of Information Technology,
Brno University of Technology,
Brno, Czech Republic
{ihurta,mrazek,drahosova,sekanina}@fit.vut.cz

**Abstract.** Taking levodopa, a drug used to treat symptoms of Parkinson's disease, is often connected with severe side effects, known as Levodopa-induced dyskinesia (LID). It can fluctuate in severity throughout the day and thus is difficult to classify during a short period of a physician's visit. A low-power wearable classifier enabling long-term and continuous LID classification would thus significantly help with LID detection and dosage adjustment. This paper deals with a co-evolutionary design of energy-efficient hardware accelerators of LID classifiers that can be implemented in wearable devices. The accelerator consists of a feature extractor and a classifier co-evolved using cartesian genetic programming (CGP). We introduce and evaluate a fast and accurate energy consumption estimation method for the target architecture of considered classifiers. The proposed energy estimation method allows for a multi-objective design enabled by introducing energy constraints. With the introduction of variable data representation bit width, the proposed method achieves a good trade-off between accuracy (AUC) and energy consumption.

## 1 Introduction

*Parkinson's disease* (PD) is one of the most common neurological conditions affecting the motor system. Patient care primarily suppresses symptoms using a *levodopa* drug, which can result in *levodopa-induced dyskinesia* (LID). A wearable device allowing long-term continuous LID classification would be a great source of information and help physicians adjust the dosage to suppress PD symptoms and, at the same time, reduce LID.

Lones et al. [4] proposed a LID-classifier model utilising a sliding window of 32 samples of low-level movement features and designed it using *genetic programming* (GP). Hurta et al. [3] further adopted this model for hardware implementation and used *cartesian GP* (CGP) [6] for the evolutionary design of an energy-efficient *feature extractor* (FE). The FE and classifier design – as a complex problem – was solved using a co-evolution approach. Their model also reduced data representation to an 8-bit integer.

The evolution design in [3] is guided only by the solution accuracy in terms of AUC (Area Under the receiver operating characteristics Curve). The number of arithmetic operations was used to estimate the hardware complexity of the final solutions. Only the best classifiers were selected and synthesised for the final evaluation. Moreover, existing works do not consider the sub-byte arithmetics, even though the sub-byte operations are currently successfully used in neural networks and machine learning accelerators [7].

To address these challenges, we propose an improved co-evolutionary design method. Our method allows for sub-byte data representation by encoding bit-width inside candidate solutions. Further, we propose an effective way of precise energy estimation during the evolution process and utilise it in a multi-objective design. Multi-objective design is achieved by transforming the multi-objective problem into a single-objective one by introducing constraints [5].

## 2    Proposed Methodology

The FE and classifier models are based on a co-evolutionary scheme proposed by Hurta et al. [3]. FE and classifier are designed simultaneously by switching the currently-evolved *population* in each epoch. Populations interact through the evaluation phase, where candidate solutions of one population are evaluated in connection with the currently best candidate solution from the other population. The *fitness* of *candidate solutions* is given as the composition's classifier accuracy (AUC). Data from the clinical study [4] is used for the fitness calculation. We also implement the co-evolution of *Adaptive Size Fitness Predictors* (ASFP) [1] that accelerates evolution.

The selection of the bit width of individual parts (i.e. FE and classifier) is incorporated into the evolution process. Bit widths are included inside the candidate solutions' chromosomes and evolve together with the cartesian grid representing the evolved program. The value of bit width spans from 3 to 12 bits and is mutated with probability equal to its portion of the chromosome.

Calculation of energy consumption traditionally involves a computationally expensive synthesis of each candidate solution. To eliminate this issue, we propose to use pre-synthesised components. Hence, we designed and synthesised each combination of 18 allowed functions and ten possible values of bit widths. This results in a look-up table of 180 different possible values of energy consumption. As the LID classifier comprises up to 32 registers, a similar table is also created for ten different bit widths of registers. Further, as classifiers generally do not utilise full sliding windows, we propose the elimination of unrequired circuits to reduce energy consumption.

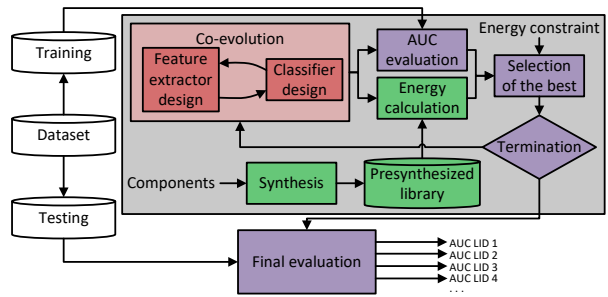

**Fig. 1.** Overview of the proposed multi-objective method for the co-evolutionary design of energy-efficient LID classifier.

The proposed solution of multi-objective design involves transforming a multi-objective problem into a single-objective. The evolution of candidate solutions is guided by the objective of classification accuracy (AUC). The objective of energy consumption is transformed into constraint $\varepsilon$. The fitness $f$ of the candidate solution (the composition of feature extractor $FE$ and classifier $C$) is equal to the achieved accuracy $AUC(FE, C)$ only if the energy consumption $E(FE, C)$ is lower than $\varepsilon$. In the opposite case, fitness $f$ is given as $-E(FE, C)$. This fitness function allows the method to improve even energy-unfit candidate solutions, which may happen especially in first generations. Combining solutions from runs with different constraints $\varepsilon$ allows for obtaining solutions spanning a wide range of energy consumption requirements while maintaining the goal of high precision and thus achieving a wide Pareto front. The overview of the proposed method is shown in Fig. 1.

## 3   Results

The adopted parameter settings are based on the settings proposed by Hurta et al. [3]. The evolutionary strategy $(1 + \lambda)$ was employed together with a limit of 10,000 generations, a grid size of 4x8 and the Goldman mutation operator [2]. One hundred independent runs of each parameter setting were performed to allow precise evaluation. Energy consumption of individual components was synthesised with Synopsys Design Compiler targeting 45 nm ASIC technology on 100 MHz frequency.

Introducing the variable width requires checking whether the maximum fitness can still be reached. For this reason, a comparison of the baseline variant with a fixed 8-bit width and the variable variant was made. Mann-Whitney U-test confirmed a non-significant difference between both variants for all test groups of the data set except LID1 and Sitting, where improvement was achieved. Modifying the initialization in the initial population from 8-bit width to a random value (in the range of 3-12) led to an additional improvement across most test groups (except for test group LID1), with a significant improvement in test groups LID3 and Siting.

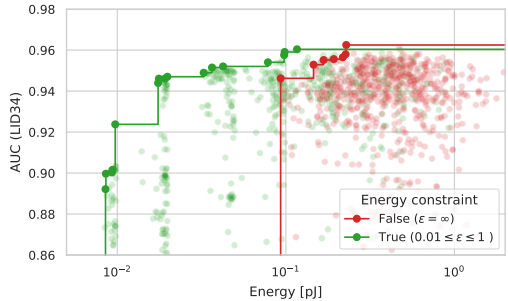

**Fig. 2.** Trade-offs between energy consumption and AUC on test group LID34. Green represents solutions obtained by a union of 100 runs for each of the seven selected energy constraints. Red represents 700 runs of the method without energy constraint.

To obtain a rich Pareto front, 100 independent runs were performed for each of seven logarithmically distributed energy constraint values from $0.01\,\mathrm{pJ}$ to $1\,\mathrm{pJ}$. To conduct a fair comparison with the standard variant, 700 independent runs with unconstrained energy consumption (i.e., $\varepsilon = \infty$) were also performed. Fig. 2 compares Pareto fronts obtained by combining the results of different energy constraint settings and the standard variant. The Pareto front achieved by the method utilizing the combined energy constraints dominates most space. In contrast, the variant without energy constraints is better only in finding a solution with the highest AUC.

## 4   Conclusions

In this paper, we proposed a method for the multi-objective design of HW accelerators for LID classifiers. The proposed efficient energy consumption estimation allowed us to include energy consumption directly into the evolution process and solve the multi-objective design problem (with a trade-off between accuracy and energy consumption) by introducing constraints on energy consumption and thus transforming it into a single-objective problem. With the introduction of variable bit width, proposed improvements allowed the design of a wide range of high-quality solutions achieving a good trade-off between accuracy and energy consumption.

**Acknowledgements** This work was supported by the Czech Science Foundation project 21-13001S and BUT Internal Grant Agency through project FIT-S-23-8141. The computational resources were supported by the MSMT project e-INFRA CZ (ID:90140). We also acknowledge Prof. Stephen Smith and Dr Michael Lones for their advice; Dr Jane Alty, Dr Jeremy Cosgrove, and Dr Stuart Jamieson for their contribution to the clinical study that generated the data, and the UK National Institute for Health Research (NIHR) for adopting the study within in its Clinical Research Network Portfolio.

## References

1. Drahosova, M., Sekanina, L., Wiglasz, M.: Adaptive fitness predictors in coevolutionary cartesian genetic programming. Evol. Comput. **27**(3), 497–523 (2019)
2. Goldman, B., Punch, W.: Reducing wasted evaluations in cartesian genetic programming. In: Krawiec, K., Moraglio, A., et al. (eds.) Genetic Programming. EuroGP 2013. LNCS, vol. 7831, pp. 61–72. Springer, Heidelberg (2013)
3. Hurta, M., Drahosova, M., Mrazek, V.: Evolutionary design of reduced precision preprocessor for levodopa-induced dyskinesia classifier. In: Rudolph, G., Kononova, A.V., et al. (eds.) Proc. of the PPSN XVII. LNCS, vol. 13398, pp. 491–504. Springer, Cham (2022)
4. Lones, M.A., Alty, J.E., et al.: A new evolutionary algorithm-based home monitoring device for Parkinson's dyskinesia. J. Med. Syst. **41**(11), 176:1–176:8 (2017)
5. Marler, R., Arora, J.: Survey of multi-objective optimization methods for engineering. Struct. Multidiscip. Optim. **26**(6), 369–395 (2004)
6. Miller, J.F.: Cartesian genetic programming. In: Cartesian genetic programming, pp. 17–34. Springer, Heidelberg (2011)
7. Mittal, S.: A survey of techniques for approximate computing. ACM computing surveys **48**(4), 1–33 (2016)

# When is an AP-solution also an ATSP-solution?

Daan van den Berg[1][0000−0001−5060−3342]

Informatics Institute
Vrije Universiteit Amsterdam
**daan@yamasan.nl**

**Abstract.** The assignment problem (AP) is very close to the asymmetric traveling salesman problem (ATSP). So close even, that some AP solutions are ATSP solutions as well, without further amendments. This is interesting because the AP is solvable in polynomial time, and ATSP is not[1]. This partial replication study shows that for 100x100 matrices, approximately 2.017% of AP solutions are also ATSP solutions, but the percentage depends on the numerical entries inside the matrix in a non-conclusive way, as the results in both studies differ on an initial segment. The community is invited to hypothesize and discuss these results, as well as directions of further study.

**Keywords:** Evolutionary Algorithms · Plant Propagation Algorithm · Metaheuristics · Parameter Control

## 1  AP & ATSP

Although the problem classes P and NP are generally believed to be fundamentally different, there are some places where they very closely touch each another. One example is the Euler cycle problem, which is to determine whether a graph has a cycle that traverses all edges exactly once, which is in P. But the almost identically formulated Hamiltonian cycle problem ("determine whether there is a cycle that traverses all *vertices* exactly once") is in NP [9,11,8]. A second example of a close touch is the Euclidean traveling salesman problem, which cannot be solved in polynomial time, but a fixed quantity error tour *can* be given in polynomial time [5,4]. A third example, and topic of this paper, is the case of the assignment problem (which is in P) and the asymmetric traveling salesman problem (which is in NP). So close are these two, that WeiXiong Zhang and Richard Korf explored the avenue of solving ATSP problem instances as AP instances [13].

Both the assignment problem and the asymmetric traveling salesman problem involve selecting exactly $n$ integers in an $n \times n$ matrix so that each row and each column contain exactly one selected integer, and the total sum of the selected integers is minimal. The matrices in Figure 1 could be assignment problem instances, with the horizontal axes denoting agents A to F and the vertical axes could denoting the jobs A to F. Note that the selected integers form actual solutions to the assignment problem: the cost to complete all jobs is minimal for all instances. But these matrices could *also* be asymmetric traveling salesman problem instances, both axes denoting cities A to F, horizontal being the city of departure, and vertical being the city of arrival, and the individual cells again the costs in minutes. The only actual difference between the two is that for an ATSP-solution, the integers need to form a single tour that includes all cities (agents, jobs) exactly once which, in this example, is only the case for the second solution.

---

[1] Assuming $P \neq NP$.

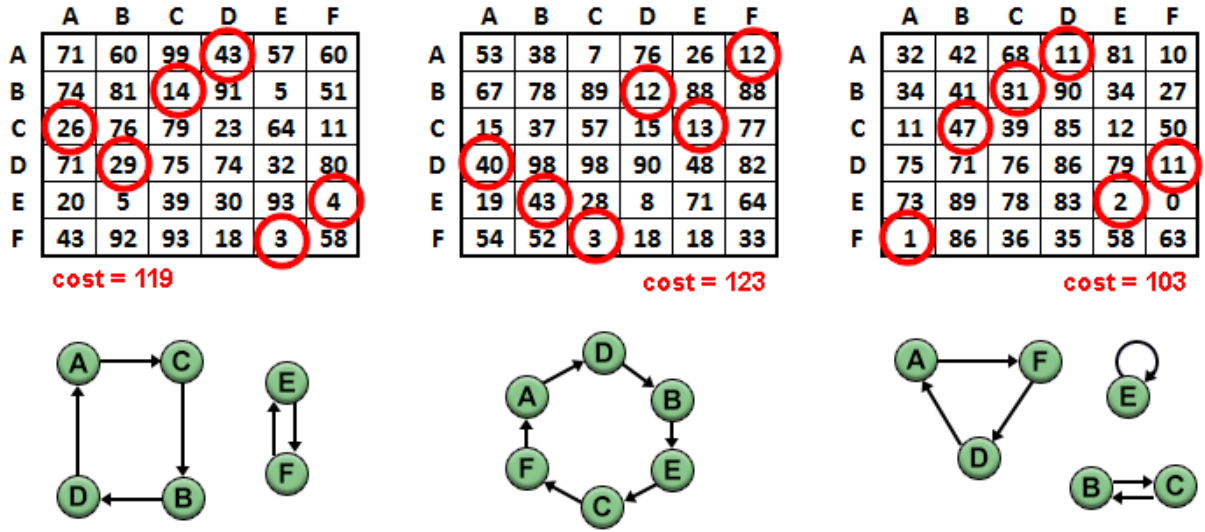

**Fig. 1.** Some assignment problem solutions (top row) are also ATSP solutions (bottom row, central graph). This interesting because in general, the assignment problem is in P, whereas the asymmetric traveling salesman problem is in NP.

But this ostensibly minuscule extra tour-requirement has immense consequences for the hardness of the problem. Whereas the assignment problem is solvable in $O(n^3)$) with the Hungarian method [6], the ATSP is NP-hard, meaning there exists no polynomial time algorithm to solve it – as yet. For this reason, it is interesting to probe the solvability boundary between the two, an endeavour that Zhang & Korf embarked on in a 1996 paper [13]. In their study, the explore whether the Hungarian method is also suitable for ATSP, else how much extra work needs to be done. But the really interesting part of their work is that they *scale* their experiment along the numerical values in the matrix, an approach that has recently seen some rekindled interest in another related problem, subset sum [1,12,7] and even ATSP itself [10]. In this paper, I will make a replication of their first results, solving the assignment problem and seeing how many solutions are also ATSP-solutions. The problem instances and source code are publicly available for further study [3].

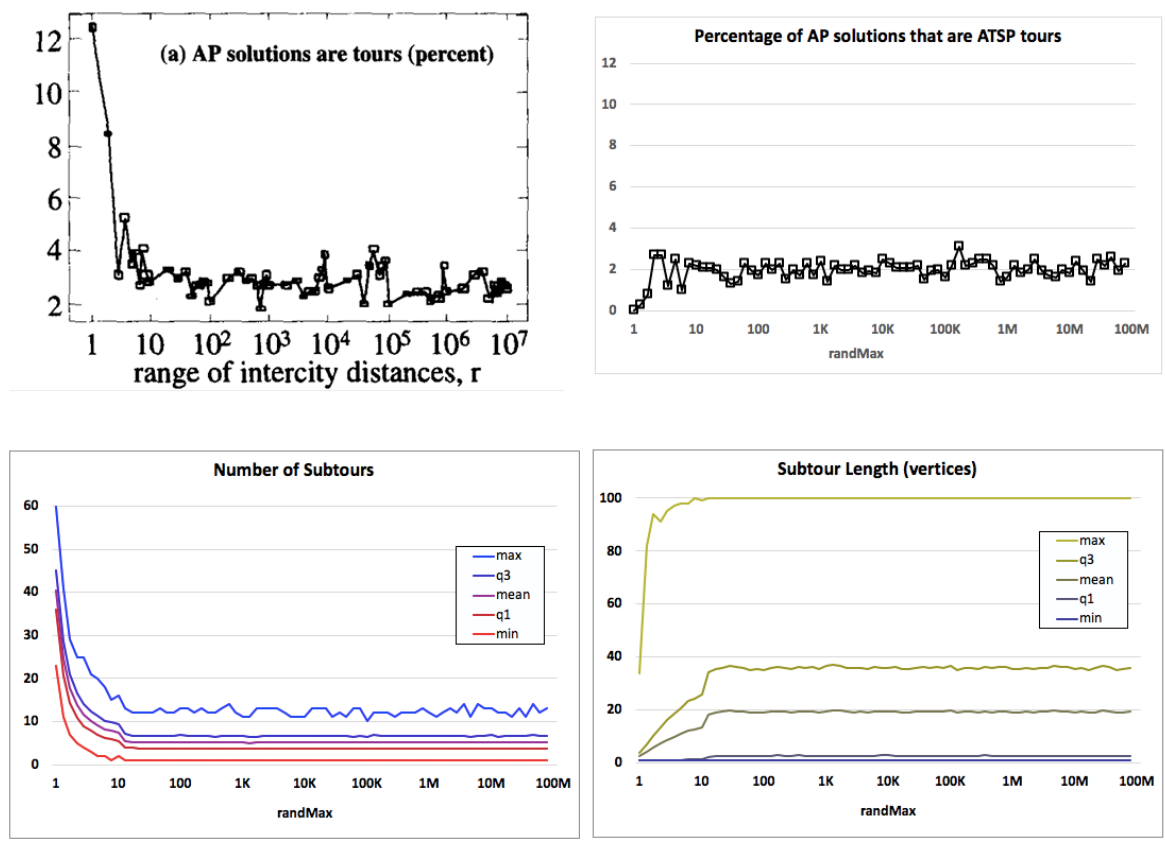

**Fig. 2.** When random matrices of 100x100 are generated with cell values over 20, about 2.07% of AP-solutions are also ATSP-solutions. The top left subfigure comes from Zhang&Korf's original paper, the top right is mine. Note how closely they resemble, except for the first 12 or so entries.

## 2   Experiment & Results

In exact correspondence to the original paper, the experiment is conducted on an ensemble of $72,000$ asymmetric matrices of 100 by 100 integers, arranged in 72 subensembles. Each subensemble has 1000 matrices which are filled with integers ranging from 0 to $randMax$, a value $\in \{1, 2, 3...10, 20, 30...100, 200, 300...90, 000, 000\}$ for each subensemble respectively. Note that after every 10th entry, the interval size increases tenfold, thus covering the entire range $1 \leq randMax \leq 90,000,000$, on a logarithmic scale. Note that these matrices *had* to be remade; 4GB is a substantial amount of data even for today's standards, let alone for 1996, and the originals do unsurprisingly not exist anymore [2].

---

[2] WeiXiong Zhang and Richard korf, personal communication, December 2022.

For each subensemble, all 1000 assignment problems are solved using the Hungarian method, which nowadays conveniently comes with Python's SciPi package [2]. In addition, I determined the number of subtours from each solution, as well as the subtour lengths. The percentage of AP solutions that are also ATSP solutions converges to $\approx 2.07\%$ for $randMax \geq 20$ which seems comparable to Zhang & Korf's results *except for the initial segment*. The same 'threshold value' of $randMax \geq 20$ appears to be applicable to the mean number of subtours, which converges to 5.18, with the maximum number of subtours converging to $\approx 12.13$ per subensemble. The mean subtour length converges to $\approx 19.31$, understandably almost the exact inverse of the mean number of subtours, with all subensembles containing at least one complete tour and one tour of length 1. For both the number of subtours and the subtour length, the averages of the first half and the last half (denoted as $q1$ and $q3$) give some idea of the distribution of the numerical results.

## 3 Conclusion

It's fair to say that the AP-to-ATSP ratio of $\approx 2.07\%$ of the solutions when $randMax \geq 20$ corresponds to the original study's results, but what happens when $randMax \leq 20$ is a mystery for now. It is unlikely due to the solving algorithm, but it should be said when $randMax \leq 20$ the expected numerical diversity of a $10,000$ cell matrix is so low that it could well host a number of equivalent minimum-cost AP-solutions, some of which might be valid ATSP tours. However, whether these supposed ATSP-solutions within these AP-solutions can also be found in polynomial time also remains an open question.

## References

1. Adriaans, P.: Differential information theory. arXiv preprint arXiv:2111.04335 (2021)
2. Anonymous: Hungarian Method in the SciPi package: https://bit.ly/3jES8qm
3. Anonymous: Repository containing source material: https://bit.ly/3ibtTiS (2022)
4. Arora, S.: Polynomial time approximation schemes for euclidean traveling salesman and other geometric problems. Journal of the ACM (JACM) **45**(5), 753–782 (1998)
5. Christofides, N.: Worst-case analysis of a new heuristic for the travelling salesman problem. Tech. rep., Carnegie-Mellon Univ Pittsburgh Pa Management Sciences Research Group (1976)
6. Kuhn, H.W.: The hungarian method for the assignment problem. Naval research logistics quarterly **2**(1-2), 83–97 (1955)
7. Sazhinov, N., Horn, R., Adriaans, P., van Den Berg, D.: The partition problem, and how the distribution of input bits affects the solving process (submitted)
8. Sleegers, J., van den Berg, D.: The hardest hamiltonian cycle problem instances: the plateau of yes and the cliff of no. SN Computer Science **3**(5), 372 (2022)
9. Sleegers, J., Berg, D.v.d.: Backtracking (the) algorithms on the hamiltonian cycle problem. arXiv preprint arXiv:2107.00314 (2021)
10. Sleegers, J., Olij, R., van Horn, G., van den Berg, D.: Where the really hard problems aren't. Operations Research Perspectives **7**, 100160 (2020)

11. Sleegers, J., Thomson, S.L., van Den Berg, D.: Universally hard hamiltonian cycle problem instances. In: ECTA 2022: 14th International Conference on Evolutionary Computation Theory and Applications. pp. 105–111. SCITEPRESS–Science and Technology Publications (2022)
12. Van Den Berg, D., Adriaans, P.: Subset sum and the distribution of information. pp. 134–140 (2021)
13. Zhang, W., Korf, R.E.: A study of complexity transitions on the asymmetric traveling salesman problem. Artificial Intelligence **81**(1-2), 223–239 (1996)

# Why is the traveling tournament problem not solved with genetic algorithms?

Kristian Verduin[1][0009−0005−8754−7635]
Thomas Weise[2][0000−0002−9687−8509]
Daan van den Berg[1][0000−0001−5060−3342]

[1]VU University Amsterdam
[2]Institute of Applied Optimization, Hefei University, China
kristian.verduin@student.uva.nl, tweise@hfuu.edu.cn, daan@yamasan.nl

**Abstract.** Because it's impossible to generate an initial population of valid individuals. When one million random instances are made, the number of constraint violations increases quadratically in instance size. Possibly, the density of feasible schedules gets extremely low for larger instances, not only prohibiting the mere creation of an initial population, but also operations such as crossover and mutation.

## 1  The Traveling Tournament Problem

It sounds so much like the Traveling Salesman Problem, that you'd be forgiven to think the Traveling Tournament Problem (TTP) is about equally hard, and could be tackled by the same metaheuristic algorithms altogether. But this is not the case. Evidence is appearing that Traveling Tournament Problem might be significantly harder, and far less amenable to traditional algorithms.

The problem is in the constraints. The TTP entails scheduling tournament rounds of an even number of (baseball) teams ($n_{teams}$) playing each other [4]. Every team plays every other team *twice* in the schedule (once home, once away), which is known as the **double round-robin** constraint. Additionally, when team A plays team B in one round, the inverse match (B playing A) cannot happen in the consecutive round, which is known as the **noRepeat** constraint. Finally, there is the maximum number of consecutive games any team can play at home or away, the **maxStreak** constraint[1]. Usually, $maxStreak = 3$ meaning any team can at most play three consecutive rounds at home, or away, anywhere in the schedule [12]. So only three constraints, but they can be violated many times per schedule, especially for larger numbers of $n_{teams}$ (Fig. 1). And as constraints go, it only takes one violation to render the entire schedule invalid.

But the actual problem is not about satisfying these constraints. The TTP, like the TSP, has a distance matrix which holds the travel time between stadiums, and the main task is to minimize the total travel time. The three constraints however, are so asphyxiating, that travel time optimization almost becomes auxiliary to finding a valid schedule in the first place.

---

[1] Terminology varies slightly across literature.

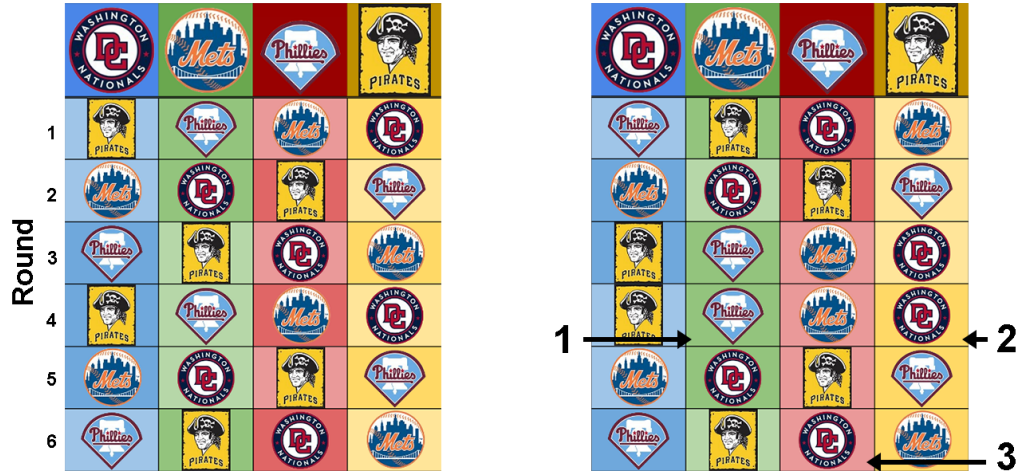

**Fig. 1. Left:** A valid TTP schedule, with lighter backgrounds designating home games, and darker away. **Right:** An invalid TTP schedule that has all three constraints violated: double round-robin (1) as the Mets play the Phillies at home *twice* in the schedule, noRepeat (2) and maxStreak (3). Note that point (1) has a noRepeat violation, in addition to the double round-robin violation.

And the problem is real. Stemming from the Major League Baseball in the United States of America and Canada, these events, and their broadcasting rights, involve vast sums of money [11]. Since its proposal in 2001, the TPP has been known as notoriously difficult to solve [6]. Proven optimal solutions have been found for classical instances with up to ten teams only. The decision variant of the problem has been proven to be NP-complete by Thielen & Westphal [12]. The optimization variant of the problem, without the inclusion of the *maxStreak* and *noRepeat* constraints, has been proven to be NP-hard [3]. Various algorithms have been studied for the TTP, including beam search [6], simulated annealing [1, 9, 8], tabu search [7], ant colony optimization [13], and integer programming [5]. However, none of these studies entail more than 16 teams, and genetic algorithms for tackling the TTP seem to be completely absent from literature. This might be due to a very surprising phenomenon: the extremely low density of feasible schedules for the problem. So low in fact, that finding an initial population in the first place might be undoable. This investigation focuses on the creation of feasible (initial) schedules and attempts to characterize the number of violations of each constraint.

## 2   Creating Random Schedules

A complete schedule for the TTP consists of $2(n - 1)$ rounds, each of which is randomly filled up by randomly selecting an opponent, after which one is designated as 'home and the other 'away'. The opponent is then marked as 'placed' and cannot be selected again in the same round.

Round by round, the entire schedule is filled up this way, which can be done in linear time. Note that although the method completely disregards the three TTP constraints (double round-robin, maxStreak, noRepeat) it is still tighter than a *complete* random fill: by using the 'augmented permutation', we ensure that every team has exactly one opponent, pretty much like a permutation in a TSP-instance ensures every sequence is a full tour. Also, teams cannot play themselves, and each match holds one home and one away team.

After filling up the schedule, the constraint violations are counted. If team A plays team B more than twice, each game after the second counts as an additional double round-robin violation for the schedule. If team A plays team B only once, this is also counted as an additional double round-robin violation. If team A does not play team B exactly once at home and once away, an additional double-round robin violation is added. If team A plays at their home or away venue more than three games consecutively, each game after the third counts as an additional maxStreak violation on the schedule. Finally, if teams A and B play each other in more than one consecutive round, each consecutive game after the first counts as an additional noRepeat violation.

For each $n_{teams} \in \{4, 6, 8...46, 48, 50\}$, one million random schedules were created, summing up to 24 million schedules. All work was done in 24 threads on SURF's Lisa Compute Cluster[2] running Debian Linux in under five hours, and our Python source code is publicly available [2].

## 3   Results and Discussion

For each $n_{teams}$, the maximum, minimum and average values for each type of constraint violation were recorded after which we fit a quadratic function to the average. On the left side of Figure 2, the filled area shows the spread of violations whereas the solid line gives the average. On the right side, the fitted quadratic functions are shown, with extrapolated values in dashed lines. Both the double round-robin violations and maxStreak violations show quadratic growth with $n_{teams}$, as $1.22x^2 - 1.69x - 0.18$ and $0.25x^2 - 0.63x$ respectively. The noRepeat violations increase linearly with $n_{teams}$ as $2.01x - 12.27$. All fits are tight, with $RMSE \leq 0.06$ for all fitted functions.

It appears that the semi-naive but really fast way of creating random initial schedules is unfeasible for schedules of any reasonable $n_{teams}$. In this experiment, the minimum number of violations over 1 million random schedules with $n_{teams} = 10$ only was 82 – nowhere *near* valid. For $n_{teams} = 20$ this number had increased to 448, after which it exploded to 1126, 2063 and 3317 for $n_{teams} = 30$, 40 and 50 respectively showing that for any reasonably sized problem instance, finding valid initial schedules with this method is practically impossible.

No wonder therefore that Anagnostopoulos et al. generate the initial schedule for their simulated annealing by backtracking by which "feasible schedules were

---

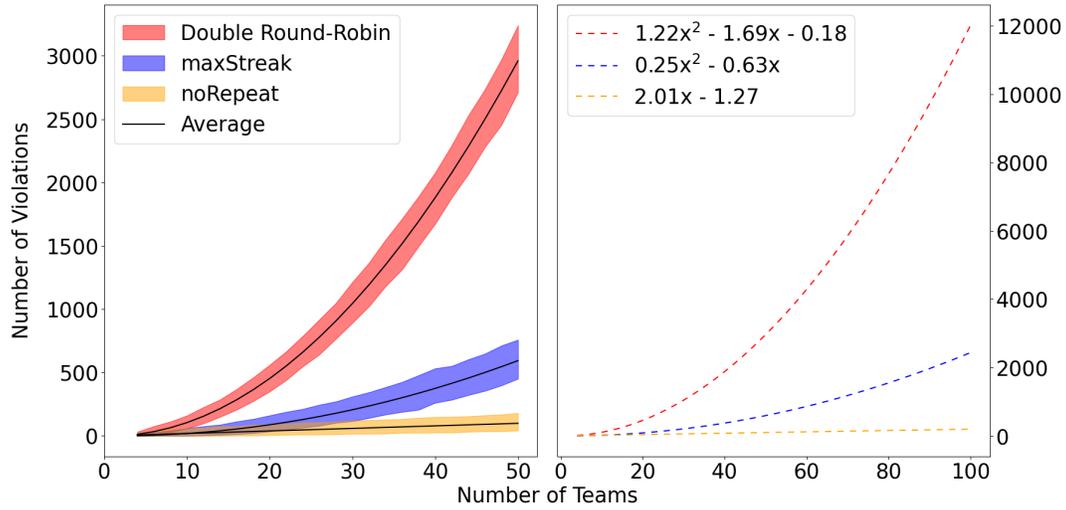

**Fig. 2. Left:** Increase in violations with the increase in the number of teams. The color-filled area is the spread of the violations (between max and min). **Right:** Quadratic fit of the violations. Dashed lines are the extrapolations of the violations.

easily obtained" [1]. But that study only uses $n_{teams} \leq 16$, and as backtracking algorithms are of exponential time complexity, much larger numbers are probably not computable. Besides, these authors treat the noRepeat and maxStreak as 'soft constraints' and their mutations might turn valid schedules into invalid ones. So even for an algorithm as simple as simulated annealing, the constraints on the TTP appear problematic.

Gaspero & Schaerf [7] and Ribeiro & Urrutia [10] both represent the schedules using one-factorization of a graph. They then continue to generate a single round-robin schedule, which is mirrored and reversed to combine into a double round-robin. It should be noted that mirrored TTP might be easier than regular TTP because its defined construction eliminates the need for the noRepeat constraint. It might also yield more expensive schedules than regular TTP, but nonetheless it is an often deployed technique in North American scheduling practice [10]. Lim et al. [9] (also) generate their initial solutions with the use of a three-phase approach. These authors also mirror their solution, and argue that doing so greatly reduces the required time for finding a feasible solution. Maybe the lesson here is that sacrificing some optimality in exchange for validity is not a bad tradeoff.

## 4   Conclusion

The average number of constraint violations for the TTP increase quadratically in the number of teams. In computer science, a polynomial increase is usually fine but remember that in this case, *all* constraints need to be satisfied before even considering a minimized travel distance – the real task at hand. It is likely for this reason that genetic algorithms are not used to solve the TTP: it's simply impossible to randomly generate an initial population of valid individuals, even before considering mutation and crossover. It serves as a stark reminder that all problems in a problem class might not be equally hard. For the TTP, being NP-hard or NP-complete might be considered a euphemism compared to the TSP, let alone to the reality of properly scheduling North American baseball teams [3].

# References

1. Anagnostopoulos, A., Michel, L., Hentenryck, P.V., Vergados, Y.: A simulated annealing approach to the traveling tournament problem. Journal of Scheduling **9**, 177–193 (2006)
2. Anonymous: Repository containing source material: https://anonymous.4open.science/r/Violation-Experiment-D56A/ (2023)
3. Bhattacharyya, R.: A note on complexity of traveling tournament problem. Optimization Online **2480** (2009)
4. Easton, K., Nemhauser, G., Trick, M.: The traveling tournament problem description and benchmarks. In: Principles and Practice of Constraint Programming—CP 2001: 7th International Conference, CP 2001 Paphos, Cyprus, November 26–December 1, 2001 Proceedings 7. pp. 580–584. Springer (2001)
5. Easton, K., Nemhauser, G., Trick, M.: Solving the travelling tournament problem: A combined integer programming and constraint programming approach. Lecture notes in computer science pp. 100–112 (2003)
6. Frohner, N., Neumann, B., Raidl, G.: A beam search approach to the traveling tournament problem. In: Evolutionary Computation in Combinatorial Optimization. pp. 67–82. Springer International Publishing (2020)
7. Gaspero, L.D., Schaerf, A.: A composite-neighborhood tabu search approach to the traveling tournament problem. Journal of Heuristics **13**, 189–207 (2007)
8. Hentenryck, P.V., Vergados, Y.: Population-based simulated annealing for traveling tournaments. In: Proceedings of the National Conference on Artificial Intelligence. vol. 22, p. 267. Menlo Park, CA; Cambridge, MA; London; AAAI Press; MIT Press; 1999 (2007)
9. Lim, A., Rodrigues, B., Zhang, X.: A simulated annealing and hill-climbing algorithm for the traveling tournament problem. European Journal of Operational Research **174**(3), 1459–1478 (2006)
10. Ribeiro, C.C., Urrutia, S.: Heuristics for the mirrored traveling tournament problem. European Journal of Operational Research **179**(3), 775–787 (2007)
11. Solberg, H.A., Gaustad, T.: International sport broadcasting: A comparison of european soccer leagues and the major north american team sports. Sport Broadcasting for Managers pp. 84–102 (2022)
12. Thielen, C., Westphal, S.: Complexity of the traveling tournament problem. Theoretical Computer Science **412**(4-5), 345–351 (2011)
13. Uthus, D.C., Riddle, P.J., Guesgen, H.W.: An ant colony optimization approach to the traveling tournament problem. In: Proceedings of the 11th Annual conference on Genetic and evolutionary computation. pp. 81–88 (2009)

---

[3] Logos in this paper were remade by melling2293@Flickr. Distributed under creative commons license.

# Stipple Tunes: An Artistic Form of Uncompressed Image in Audio Steganography

Alexa Lewis[1], Christopher J Tralie[1][0000−0003−4206−1963]

Ursinus College, Department of Mathematics And Computer Science, Collegeville, PA, USA

**Abstract.** We present an artistic audio steganography technique for hiding stipple images inside of uncompressed audio that we dub "stipple tunes." Given an audio carrier and a stipple image to hide, the goal is to manipulate samples in the left and right audio channels to draw the stipple points; that is, the left and right channels are interpreted, respectively, as X and Y coordinates in the Cartesian plane. To accomplish this, we devise an objective function that pans the audio and restricts samples to the stipple, while minimizing error, which we solve using the Viterbi algorithm. Decoding the hidden image is trivial; we simply create a scatterplot of the audio samples. We provide code, examples, and an interactive viewer in Javascript at `https://ctralie.github.io/StippleTunes/Viewer/`

## 1 Introduction

Steganography is the process of hiding one data stream "in plain sight" in another "carrier" data stream. In audio steganography [6][8], audio acts as a carrier. In any steganography technique, a simple scheme involves hiding data in the least significant bit of samples [5]. In 16-bit audio, this is inaudible. The downside of such techniques is that compression will destroy the hidden data, and it is easy to detect statistically [9]. However, in this work, we treat steganography as more of an artistic endeavor; we are concerned less with the data being compressed or statistically or audibly hidden, as long as the audio is still pleasant to listen to. To that end, we pursue an uncompressed audio steganography technique that we dub "stipple tunes," which is specifically designed to hide images in audio. Our goal is to spread an audio carrier across two channels in such a way that each pair of audio samples, when thought of as a point, plots an XY scatterplot that creates the hidden image. Figure 1 and Figure 2 show two examples.

Using the audio channels as coordinates on the Cartesian plane is spiritually similar to oscilloscope music [1][2][14], though we don't "connect the dots" as an oscilloscope would. It is also worth mentioning recent works that trained neural networks to hide full resolution color images [4][10][13][7] in audio, though we want our technique to be easier to explain and implement.

## 2    Stipple Tunes

To hide images in audio, we first turn to an intermediate representation: the *stipple pattern*, or a collection of dots that resembles the image. We use the technique of Secord [12] to automatically create stipples. This technique samples randomly from a density function that is higher in darker regions of the image, and then it moves the dots towards their Voronoi centers repeatedly (Lloyd's algorithm) until they converge to a more uniform, aesthetically pleasing distribution. To make sure our image picks up on important edges, even if they are brighter, we also make the density function higher in regions that are closer to edges, which we detect with a Canny edge detector [3].

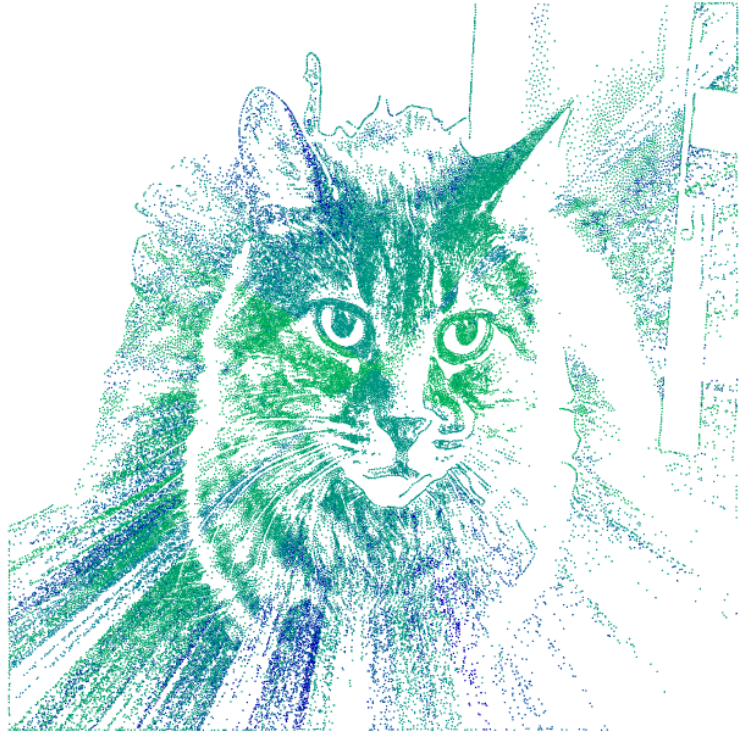

**Fig. 1.** A stipple tune on Layla the cat, using a 30 second clip from Eric Clapton's "Layla," created from a stipple with 100,000 points.

Once we have the stipple pattern and a single channel audio carrier $x[j]$, we turn the audio stream into a 2D curve by simply repeating the channel twice: one for each coordinate. From there, a simple idea is to find the nearest neighbor in the stipple pattern to each 2D audio point $(x[j], x[j])$. However, this has an immediate drawback since the curve simply moves back and forth along the line $y = x$, so nearest neighbors would concentrate near this line. To encourage the algorithm to explore points away from this line, we slowly rotate the line and sweep the entire stipple. This corresponds to panning the audio between two channels which, while audible, is not unpleasant.

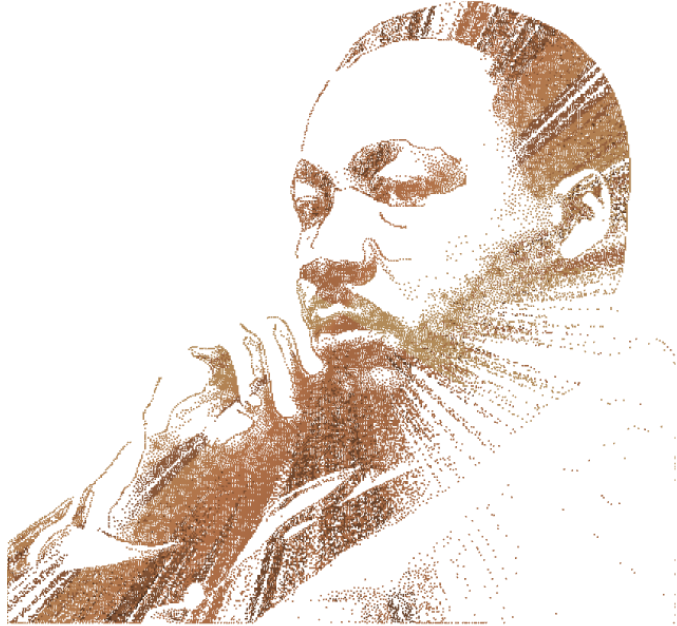

**Fig. 2.** A stipple tune on a picture of Martin Luther King Jr., using a 30 second clip from his "I Have A Dream" speech.

Formally, let $Y[j] = x[j] + ix[j]$ be an embedding of $(x[j], x[j])$ in the complex plane, and similarly embed the stipple $Z$ in the complex plane. Then, we introduce a hidden state $\theta[j]$ so that we actually find the nearest neighbor from the points $Y_\theta[j] = Y[j]e^{i\theta[j]}$ to the stipple pattern $Z$. The effect of $\theta[j]$ is to pan between the left and right audio channels, and snapping $Y_\theta[j]$ to the nearest point in the stipple can be thought of as an unusual form of quantization.

Crucially, we encourage the line to move and sweep the whole image by forcing $\theta[j+1] > \theta[j] + \epsilon$ for some $\epsilon > 0$. We can solve for the hidden states $\theta[j]$ using the Viterbi algorithm. Rather than maximizing a probability, as in the traditional application of Viterbi to HMMs, we seek to *minimize* the sum of nearest neighbor distances[1]. In this way, our application is similar in spirit to corpus-based concatenative synthesis [11], where the "corpus" is simply the stipple pattern. Algorithm 1 provides more details. In practice, we discretize $\theta$ by a factor *win* coarser than audio sample rate to keep the Viterbi algorithm tractable. We also discretize the possible rotation angles to *na*, and we force

---

[1] Of course, a probability can always be converted to a "distance" via a negative log

adjacent angle states to be between 1 and $tw < na$ of each other so that adjacent angles have to change, but not by an arbitrary amount. We also use a grid (with resolution equal to that of the image) to perform approximate nearest neighbor of points in $Y$ to stipple points $Z$.

Figure 1 shows an example of mapping a stipple of a special cat named Layla to a 30 second clip from Eric Clapton's "Layla," using $na = 60$, $win = f_s = 44100$, and $tw = 10$. Since $win$ is the sample rate $f_s$, we only have one state per second, but we find this is enough to get a good sweep through the stipple.

---

**Algorithm 1** Stipple Tunes Algorithm

---

1: **procedure** STIPPLETUNE($Z$, $x$, na, win, tw)          ▷ $Z$ is stipple, $x$ is audio
   samples, $na$ is number of discrete angle states, $win$ is number of samples between
   angle states, and $tw$ is amount by which angle can jump each step
2:     $N \leftarrow \text{len}(x)$                          ▷ Number of audio samples
3:     $Y \leftarrow x + ix$
4:     $M \leftarrow \text{ceil}(N/\text{win})$
5:     $C[i,0] \leftarrow 0, C[i, j > 1] \leftarrow \infty$          ▷ $na \times M$ Cumulative cost matrix
6:     $I[i,j] \leftarrow 0$                          ▷ $na \times M$ backpointers to best preceding state
7:     **for** $t = 2 : M$ **do**
8:         **for** $j = 1 : na$ **do**
9:             $\theta_j \leftarrow 2\pi j/\text{na}$
10:            **for** $k = j - \text{tw} : j - 1 \mod na$ **do**
11:                $\theta_k \leftarrow 2\pi k/\text{na}$
12:                Let $\theta_\ell \leftarrow \theta_j + (\theta_k - \theta_j)/\text{win}$
13:                $d \leftarrow \sum_{\ell=1}^{\text{win}} dZ_{NN}(Y[\text{win} * t + \ell]e^{i\theta_\ell})$          ▷ Sum distances to the nearest
                   neighbors in $Z$ of all rotated $Y$ points
14:                **if** $C[k, t-1] + d < C[j, t]$ **then**
15:                    $C[j,t] \leftarrow C[k, t-1] + d$
16:                    $I[j,t] \leftarrow k$                          ▷ Remember optimal transition
17:                **end if**
18:            **end for**
19:        **end for**
20:    **end for**
21:
22:    Backtrace $I$ to obtain the optimal sequence of angle states
23:    Linearly interpolate between each angle state (line 11) to compute $\theta[k], k = 1$ to $N$
24:    Let $X_k$ be the nearest neighbor in $Z$ to $Y_k e^{i\theta[k]}$
25: **return** $X$
26: **end procedure**

---

# 3  Comparison To LSB Steganography

Let's suppose our audio is sampled at 44100hz. Then an LSB technique on 2 channel audio would transmit 88200 bits/second. Let's further suppose our stipple has been discretized to a 1024x1024 grid, so that each stipple location requires 20 bits to transmit. This means that we could transmit 4410 stipple locations per second with an LSB technique, or 132,300 samples over a 30 second span. By contrast, with our technique, we could technically get a stipple location at every sample, for 10x as many locations per second. However, it is unlikely that Algorithm 1 would choose every stipple point if we used this many. In both Figure 1 and Figure 2, we used stipples with 100,000 samples over a period of 30 seconds, and most stipple samples were chosen, so this has a similar capacity to the LSB technique. However, our technique is incredibly easy to decode; we simply create a scatterplot of the 2 channel audio samples. This is also easier to explain to a non technical audience than LSB encoding.

## References

1. Rendering shapes through audio signals. `https://felixonline.co.uk/issue/1773/science/rendering-shapes-through-audio-signals`, accessed: 2022-07-19
2. Ball, J.H.: Osci-render. `https://github.com/jameshball/osci-render` (2022), accessed: 2022-07-19
3. Canny, J.: A computational approach to edge detection. IEEE Transactions on pattern analysis and machine intelligence (6), 679–698 (1986)
4. Cui, W., Liu, S., Jiang, F., Liu, Y., Zhao, D.: Multi-stage residual hiding for image-into-audio steganography
5. Cvejic, N., Seppanen, T.: A wavelet domain LSB insertion algorithm for high capacity audio steganography. In: Proceedings of 2002 IEEE 10th Digital Signal Processing Workshop, 2002 and the 2nd Signal Processing Education Workshop. pp. 53–55. IEEE
6. Djebbar, F., Ayad, B., Meraim, K.A., Hamam, H.: Comparative study of digital audio steganography techniques **2012**(1), 25
7. Domènech Abelló, T.: Hiding images in their spoken narratives. Master's thesis, Universitat Politècnica de Catalunya (2022)
8. Dutta, H., Das, R.K., Nandi, S., Prasanna, S.R.M.: An overview of digital audio steganography **37**(6), 632–650
9. Fridrich, J., Goljan, M., Du, R.: Reliable detection of lsb steganography in color and grayscale images. In: Proceedings of the 2001 workshop on Multimedia and security: new challenges. pp. 27–30 (2001)
10. Geleta, M., Punti, C., McGuinness, K., Pons, J., Canton, C., Giro-i Nieto, X.: PixInWav: Residual steganography for hiding pixels in audio
11. Schwarz, D.: Corpus-based concatenative synthesis. IEEE signal processing magazine **24**(2), 92–104 (2007)
12. Secord, A.: Weighted voronoi stippling. In: Proceedings of the 2nd international symposium on Non-photorealistic animation and rendering. pp. 37–43 (2002)
13. Takahashi, N., Singh, M.K., Mitsufuji, Y.: Source mixing and separation robust audio steganography
14. Teschler, L.: Making pictures from sound on an oscilloscope. `https://www.testandmeasurementtips.com/making-pictures-from-sound-on-an-oscilloscope-faq/`, accessed: 2022-12-01

# Enhancing an Intrusion Detection System by means of a  Genetic Algorithm[⋆]

A.M. Mora[1], P. Merino[1], D. Hernández[1]

Departamento de Teoría de la Señal, Telemática y Comunicaciones.
ETSIIT-CITIC, Universidad de Granada.
`amorag@ugr.es,pablomerino25@gmail.com,diegohm8991@gmail.com`

## 1 Introduction

Cybersecurity is a critical issue due to the exponential growth of network traffic and the increasing focus of threats on computer networks. Cyberattacks can have severe consequences, including unauthorized access, destruction, modification, disclosure of information, or denial of services, and they must be prevented or detected as soon as possible. Effective Intrusion Detection Systems (IDSs) [1] are a crucial defense method against hostile activity on a network. IDSs can be either Host-based (HIDSs), monitoring critical hosts in the network, or Network-based (NIDSs), focused on monitoring data transported by the network. NIDSs are algorithms that can identify malicious or suspicious patterns inside network traffic analyzed in real-time, and they belong to two main types: sign-based detectors and anomaly-based detectors.

MSNM [2] is a state-of-the-art anomaly-based IDS that applies Multivariate Statistical Network Monitoring to detect anomalies inside a huge dataset of network traffic flows. It works well for a large number of input variables but has a limitation, as it values all input variables equally, making it difficult to detect attacks that alter only some of these variables. To deal with this problem, the authors of MSNM developed a semi-supervised version that assigns different weights to each variable in order to set its importance in the detection. However, optimizing these weights for each type of attack is a challenge.

This study aims to improve the results obtained by MSNM IDS by enhancing the optimization process of the variable weights. Different Evolutionary Algorithms (EAs) [3] variations will be designed and applied. EAs are a family of metaheuristics inspired by natural evolution. The study describes the adapted approaches regarding their codification, genetic operators, and fitness functions. The methods are tested in different experiments using the UGR'16 dataset [4] to compare the results with those obtained by the standard MSNM method.

---

[⋆] This work has been partially funded by projects PID2020-113462RB-I00 (ANIMALICOS), granted by Ministerio Español de Economía y Competitividad MCIN/AEI/10.13039/501100011033 and European Union NextGenerationEU/PRTR; projects P18-RT-4830 and A-TIC-608-UGR20 granted by Junta de Andalucía, and project B-TIC-402-UGR18 (FEDER and Junta de Andalucía)

In conclusion, cyberattacks are a serious threat to computer networks and must be prevented or detected as soon as possible. IDSs are a crucial defense method, and MSNM is a state-of-the-art anomaly-based IDS that detects anomalies inside a huge dataset of network traffic flows. However, optimizing the variable weights is a challenge, and this study aims to improve the results obtained by MSNM IDS by enhancing the optimization process of the variable weights using different EAs variations and PSO.

## 2   UGR'16

This dataset can be accessed and downloaded at **https://nesg.ugr.es/nesg-ugr16/**. It comprises network flow traces obtained by monitoring traffic within an actual Internet Service Provider (ISP) network over a span of five months. The data gathering occurred in two phases, employing Netflow sensors. The initial collection was conducted from March to June of 2016, encompassing typical network usage scenarios. The objective was to scrutinize and model the standard behavior of network users while detecting irregularities such as SPAM campaigns. Subsequently, the flows were categorized as either 'background' (representing legitimate flows) or 'anomalies' (indicating non-legitimate flows), serving the purpose of training models in the CALIBRATION segment of the dataset.

The second phase unfolded between July and August of 2016, during which 'controlled' (synthetic) attacks were executed to generate a TEST dataset for validating anomaly detection algorithms. Twenty-five virtual machines were deployed within a sub-network of the ISP, with five of these machines launching attacks against the remaining twenty. This segment constitutes the TEST section of the dataset, utilized for validating the trained models.

For this preliminary work, we have considered a subset of the whole dataset, since it has more than 200GB of data. Thus, there are two weeks in the month of July, during which there were normal/background traffic and also synthetic attacks.

There are 4 different types of attacks labelled: DOS (Denial of Service), SCAN44 (Port Scan with 4 sources and 4 victims), SCAN11 (Port Scan with 1 source and 1 victim) and NERIS (Neris Botnet attack).

## 3   Methods

Multivariate Statistical Network Monitoring (MSNM) was proposed by Camacho et al. [2]. It is an IDS based on anomaly detection, which depends on 134 weights (the most influential variables). The same authors implemented a method to set the weights optimally, named Run-to-Run PLS (R2R-PLS) [5], being Partial Least Squares (PLS) optimization a multivariate regression technique, that makes MSNM a semi-supervised learning method.

In the present study, we propose some Evolutionary Algorithms (with different configurations) to optimize the weights, comparing the obtained results with R2R-PLS approach.

The CALIBRATION segment of the UGR'16 dataset was employed to refine the candidate solutions of the algorithms and adapt the MSNM accordingly. On the other hand, the TEST portion was utilized to assess the optimal final individual or solution in each instance, without incorporating this data for fitness computation during optimization.

The proposed EA is a Genetic Algorithm [6], in which every individual is a vector of 134 values (weights), each of them in [0,1]. The aim is to maximize the fitness function, which is computed in base of the attack detection performance. The population is initialized randomly. We considered generational and steady-state approaches, so half of the population is selected as parents for the following generation or just two of them, respectively. Two different selection mechanisms are implemented: a random selection of parents and a selection based on their fitness, where the best individuals are chosen. With regard to the crossover, again two different operators have been implemented: a uniform and a two-point crossover. There is a random mutation operator, applied (depending on the mutation probability) to a number of genes of the generated individuals. This number depends on the so-called mutation rate, that is a percentage of the whole chromosome (between 0 and 1). Finally, the replacement policy considered, potentially substitutes half of the population with the generated offspring, however, just those that have a worse fitness will be replaced by a new individual.

## 4  Experiments and Results

The experiments were conducted utilizing the Global Optimization Toolbox (gOpt) [7] and the MEDA Toolbox [8] within MATLAB. Given the stochastic nature of the algorithms employed, 10 runs were executed for each configuration, with results averaged and standard deviations calculated accordingly. A distinct set of runs was performed for each type of attack, as the optimal weight sets are heavily reliant on attack type.

The primary metric utilized for algorithm evaluation, as well as the fitness function for every individual within the Evolutionary Algorithms (EAs), is the Area Under the Curve (AUC) [9], as obtained by semi-MSNM using optimized weights. AUC is associated with the Receiver Operating Characteristic (ROC) curve, which depicts the true positive rate versus false positive rate in a classification system. Consequently, AUC yields a value between 0 and 1, where a value of 1 signifies a perfect solution, indicating that the optimized MSNM has successfully detected all anomalies within the dataset. Conversely, a value of 0 suggests no anomalies were detected, while a value of 0.5 implies random classification.

Thus, the fitness of an individual in the GAs is the value of the aforementioned AUC after applying MSNM method considering the corresponding set of weights for the variables that the individual has.

In the first experiment we run the GA 10 times trying to detect each of the 4 types of attacks. We then analyze the evolution of the fitness values in each case. The obtained results in one example run are presented in Figure 1.

**Fig. 1.** Progression of the best (maximum) and average fitness across various runs of the algorithm. Each type of attack is represented by an individual run.

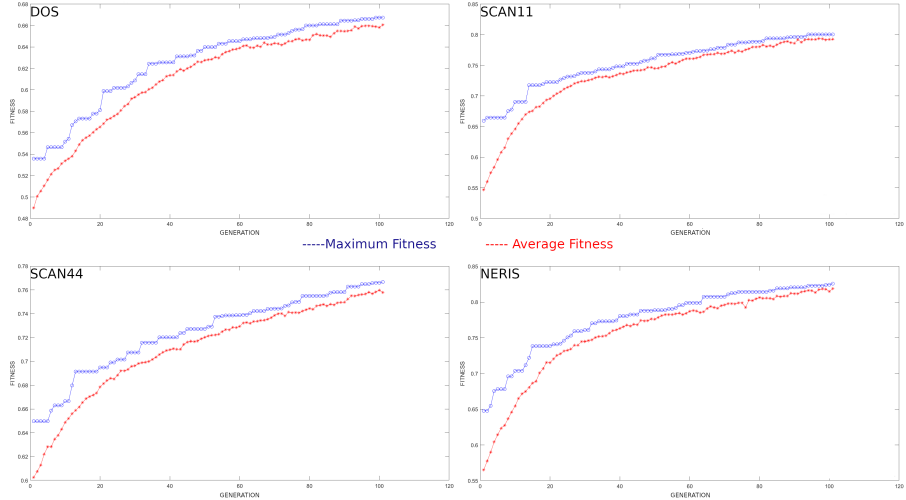

The figure illustrates that there is a consistent trend toward maximization in both the maximum and average fitness of the population, albeit with some anticipated fluctuations in the average, as observed in the SCAN44 run. Given that Genetic Algorithms (GAs) are inherently stochastic methods, it's typical for certain runs to commence with a less favorable population, necessitating several generations, or possibly never reaching a stable state in the quality of individuals within the population. Nevertheless, these findings suggest that the algorithm is functioning effectively.

Once, we have tested that the GA behaves properly with regard to the fitness improvement, we test the algorithm for optimizing the weights for the Semi-supervised MSNM method. We tested first a Steady-State approach with the following configuration: 40 individuals, 500 generations, crossover probability 1, mutation probability 0.5, mutation rate 0.02.

After thorough testing, we applied the GA to optimize the weights for the Semi-supervised MSNM technique. Initially, we employed a Steady-State approach with the following parameters: 40 individuals, 500 generations, crossover probability set to 1, mutation probability of 0.5, and a mutation rate of 0.02. Figure 2 shows the obtained results. In it, GENETIC is the GA model, MSNM-AS is the first version of the IDS, and MSNM-R2R-PLS is the optimized version with their own method.

The superiority of the GA results over the initial implementation of MSNM, which solely conducts auto-scaling on variables, is evident, except in the case of the SCAN11 attack. Additionally, improvements are observed in certain scenarios compared to the optimized version MSNM-R2R-PLS. However, the genetic approach exhibits notably poor performance in handling NERISBOTNET compared to the semi-supervised MSNM. This observation prompts us to reconsider whether the Steady-State method is the most suitable approach for addressing this particular problem.

**Fig. 2.** Results obtained for the GA approach (GENETIC) considering a Steady-State model, 10 runs.

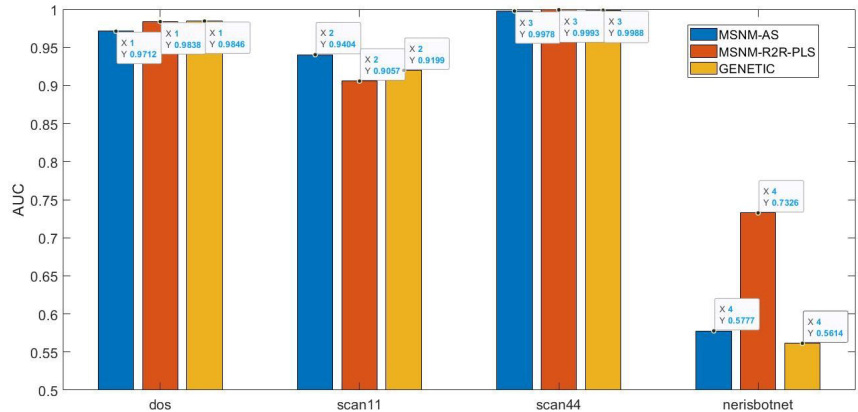

Thus, in the following experiment, we will test a Generational model of the GA. The considered configuration is: 40 individuals, 100 generations, crossover probability 1, mutation probability 0.5, mutation rate 5. Figure 3 illustrates the results obtained (averaged over 10 runs), focusing on various types of attacks. It is evident that the Generational model achieves a 5% enhancement in detecting NERISBOTNET, addressing the deficiency observed in the Steady-State approach. However, the results for DOS and SCAN11 attacks are slightly inferior compared to other methods. This suggests that while this approach demonstrates overall improvement, there remains potential for enhancement, particularly in accurately detecting NERISBOTNET traffic flows, which pose a significant challenge even for the Semi-supervised MSNM IDS.

## 5 Conclusions and future work

This work presents a preliminary study on the application of different schemes of Genetic Algorithms to optimize an Intrusion Detection System (IDS) named Multivariate Statistical Network Monitoring. It is based on the use of a set of weights in order to detect anomalies (potential cyberattacks) in a dataset of real network flows.

We have applied two different Genetic Algorithm approaches, a generational and a steady-state. The obtained results show that the application of GAs can improve the detection ability of the IDS with regard to three of the four attacks to detect.

**Fig. 3.** Results obtained for the GA approach (GENETIC) considering a Generational model, 10 runs.

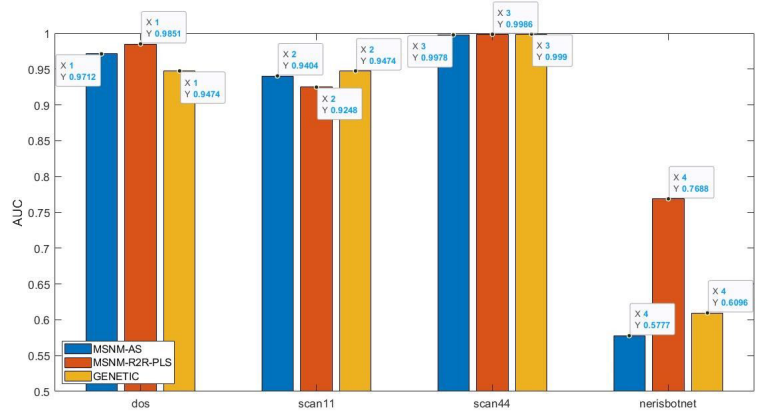

However, there is still room for improvement with respect to the botnet attack (NERISBOTNET), in which we will mainly focus in the following steps of this research.

## References

1. Sabahi, F., Movaghar, A.: Intrusion detection: A survey. In: 2008 Third International Conference on Systems and Networks Communications, IEEE (2008) 23–26
2. Camacho, J., Pérez-Villegas, A., García-Teodoro, P., Maciá-Fernández, G.: Pca-based multivariate statistical network monitoring for anomaly detection. Computers & Security **59** (2016) 118–137
3. Eiben, A.E., Smith, J.E.: Introduction to evolutionary computing. Springer (2015)
4. Maciá-Fernández, G., Camacho, J., Magán-Carrión, R., García-Teodoro, P., Therón, R.: Ugr '16: A new dataset for the evaluation of cyclostationarity-based network idss. Computers & Security **73** (2018) 411–424
5. Camacho, J., Picó, J., Ferrer, A.: Self-tuning run to run optimization of fed-batch processes using unfold-pls. AIChE journal **53**(7) (2007) 1789–1804
6. Goldberg, D.E.: Genetic Algorithms in search, optimization and machine learning. Addison Wesley (1989)
7. Lofberg, J.: Yalmip: A toolbox for modeling and optimization in matlab. In: 2004 IEEE international conference on robotics and automation (IEEE Cat. No. 04CH37508), IEEE (2004) 284–289
8. Camacho, J., Pérez-Villegas, A., Rodríguez-Gómez, R.A., Jiménez-Mañas, E.: Multivariate exploratory data analysis (meda) toolbox for matlab. Chemometrics and Intelligent Laboratory Systems **143** (2015) 49–57
9. Hastie, T., Tibshirani, R., Friedman, J.H., Friedman, J.H.: The elements of statistical learning: data mining, inference, and prediction. Volume 2. Springer (2009)

# Population size influence on the energy consumption of Genetic Programming

Josefa Díaz-Álvarez[1],0000-0003-2105-3905, Maribel G. Arenas[2],0000-0002-5576-4417, Pedro A. Castillo[2],0000-0002-5258-0620, Francisco Fernández de Vega[1],0000-0002-1086-1483, Francisco Chávez[3],0000-0002-0691-292X and Jorge Alvarado[4]0000-0002-7943-4455

Computer architecture department, University of Extremadura. C. Santa Teresa de Jornet, 38, 06800, Mérida, Spain. {mjdiaz, fcofdez}@unex.es
Department of Computer Engineering, Automation and Robotics. ETSIIT and CITIC, University of Granada, Granada, Spain. {mgarenas, pacv}@ugr.es
Computer and Telematics Systems Department, University of Extremadura. C. Santa Teresa de Jornet, 38, 06800, Mérida, Spain. fchavez@unex.es
GEA group, University of Extremadura. C. Santa Teresa de Jornet, 38, 06800, Mérida, Spain. jalvarado@unex.es

**Abstract.** Evolutionary Algorithms (EAs) are routinely applied to solve a large set of optimization problems. Traditionally, their performance in solving those problems is analyzed using the fitness quality and computing time, and the effect of evolutionary operators on both metrics is frequently analyzed to compare different versions of EAs. Nevertheless, scientists face nowadays the challenge of considering energy efficiency in addition to computational time, which requires studying the algorithms energy consumption.

This paper discusses the interest in introducing *power consumption* as a new metric to analyze the performance of genetic programming (GP). Two well-studied benchmark problems are addressed on three different computing platforms, and two different methods to measuring the power consumption have been tested.

Analyzing the population size, the results demonstrate its influence on the energy consumed: a non-linear relationship was found between the size and energy required to complete an experiment. This study shows that not only computing time or solution quality must be analyzed, but also the energy required to find a solution.

Summarizing, this paper shows that when GP is applied, specific considerations on how to select parameter values must be taken into account if the goal is to obtain solutions while searching for energy efficiency. Although the study has been performed using GP, we foresee that it could be similarly extended to EAs.

Energy consumption, evolutionary algorithms, energy-aware computing, performance measurements.

# 1 Introduction

Energy efficiency has acquired remarkable importance in the last decades in the whole field of Information Technology (IT), which affect not only low level design -such as circuit level, cache memory resizing, or refreshing, but also large IT infrastructures design, including cooling systems required, as well as the design of algorithms considering code optimizations, programming language selection and paremeterization. This last issue is the one we address in the paper recently published in [2], and of which we present here a summary.

Considering the energy consumption of algorithms, several proposals have already been published, such as the one in 2015 that studied power consumption for several sorting algorithms with different programming languages and architectures. A comparative analysis using symmetric and asymmetric encryption algorithms was published in 2017; also in 2018, a routing protocol based on a genetic algorithm for a middle-layer oriented network consuming less energy than the previous one was presented.

As described above, some algorithms have already been analyzed from the point of view of power consumption. But in the case of EAs, this idea has only been presented in the literature recently, and it needs to be researched as Camacho et al. proposed in [1]. For these new approaches, the power consumption metric should be included as result, making, in the end, more efficient algorithms. For the moment our contribution is focused on the study of algorithm parameters, such as population size; execution time, or the problem complexity, on standard Genetic Programming (GP), using some benchmark problems where we have concluded the relation between parameters and energy consumed is not as linear as we could have expected. So, the reasons for that behavior need to be addressed in the future more in detail checking if that is true for other EAs.

The rest of the paper is organized as follows: in section 2 we describe previous approaches in the area. Section 3 describes the methodology for the analysis and Section 4 presents the results and conclusions obtained.

# 2 Power Consumption and EAs

During the last few years, the power consumption metric has been included in some research papers making energy-efficiency one of the main goals of the algorithms. Detailed information can be found in [2].

Closer to our goals, [3] reviewed different methods and power models to estimate power consumption in machine learning and we have used some of them for our work, also with additional measurement procedures.

Other parameters, like population and chromosome size, were studied for the first time in [4], having in mind the energy consumption patterns, but the conclusions did not include GP algorithms.

Related to the experiments we take on the McDermott suggestions and have selected two classical GP problems for bechmarking, *Lid* and *Order Tree*. With these problems, we can adjust the problem size - and thus its complexity - and calculate the energy consumption of the GP algorithm.

## 3 Methodology

This work hypothesized that the main EA parameters not only influence computing time or quality of solutions, but they highly influence the total energy consumed by the algorithm when addressing a problem.

It is specifically applied to GP and one of its main parameters: population size. Experiments were carried out on ARM and x86 architectures on three different computer devices (Raspberry pi, laptop, and PC), where GP benchmarks (*LID* and *ORDER Tree*) are launched using ECJ. We used a digital power meter (*Yokogawa WT310E*) and *APPPowerMeter* software application as instrumentation tools to measure energy consumption. Detailed information can be found in [2].

## 4 Results and Conclusions

Three different experiment settings were carried out for *LID* and *ORDER* on the different devices, which are summarized as follows:

– Experiment #1: Running experiments for 300 seconds, and analyzing energy consumption for every population size considered.
– Experiment #2: Running experiments until finding solution.
– Further Experiments: We conduct further experiments based on the conclusions of the two previous ones.

We show some details about the *Experiment #1* of the x86 architecture, a PC platform. Readers can find the whole set of experiments and results in [2]. Fig. 1 (a) and (b) represents the growth rates of energy consumption and runtime, between consecutive population sizes for *LID* and *ORDER Tree*, respectively. Table 1 describes the evolution of fitness values and a number of generations computed for each device.

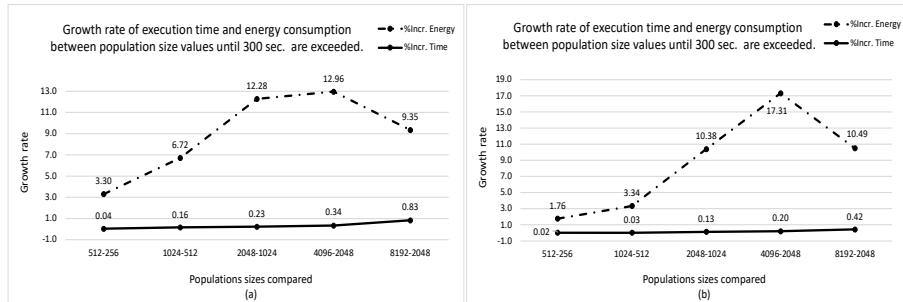

**Fig. 1.** Analyzing the growth rates of execution time vs power consumption for each population size on the PC platform. Graph (a) corresponds to *LID* and (b) to *ORDER Tree*

**Table 1.** Summary of the average best fitness values for each population size for 30 runs and both *Lid* and *ORDER Tree* problems. *Fv* corresponds to the fitness value and *Ge* represents the average number of generations.

| Evolution in fitness value for experiments with time limit 300 seconds. | | | | | | | | | | | | |
|---|---|---|---|---|---|---|---|---|---|---|---|---|
| | Raspberry Pi | | | | Laptop | | | | PC | | | |
| Pop. Size | Lid | | ORDER | | Lid | | ORDER | | Lid | | ORDER | |
| | Fv | Ge | Fv | Ge | Fv | Ge | Fv | Ge | Fv | Ge | Fv | Ge |
| 256 | 50.08 | 52.67 | 229.27 | 124.50 | 58.33 | 8708.53 | 579.07 | 4883.93 | 48.56 | 7499.70 | 709.47 | 4953.80 |
| 512 | 47.96 | 30.70 | 201.67 | 81.27 | 55.57 | 4041.83 | 793.73 | 2137.43 | 55.57 | 4861.40 | 694.63 | 2519.60 |
| 1024 | 40.57 | 18.33 | 179.30 | 53.13 | 53.23 | 1794.37 | 909.40 | 912.60 | 53.25 | 2195.33 | 902.47 | 1339.83 |
| 2048 | 30.55 | 13.00 | 129.53 | 35.57 | 55.57 | 936.27 | 756.73 | 442.23 | 53.23 | 1076.53 | 959.47 | 701.50 |
| 4096 | 29.85 | 10.80 | 103.23 | 24.30 | 41.59 | 230.63 | 627.47 | 276.70 | 41.54 | 312.77 | 802.13 | 387.10 |
| 8192 | - | - | - | - | 36.87 | 65.90 | 493.63 | 155.73 | 41.54 | 151.30 | 609.10 | 211.30 |

Although more experiments in the future with a wider set of problems will be useful to confirm what we see here, we can draw some results conclusions of interest: Firstly, population size influences energy consumed, but there is not a linear relationship between size and energy required to complete the experiment. Secondly, the complete experiments showed that the energy consumption is not directly correlated with runtime. All in all, this paper paves the way for future research on the influence and proper selection of GP and EA main parameters when looking for energy-efficient algorithms.

## Acknowledgment

We acknowledge support from Spanish Ministry of Economy and Competitiveness under projects PID2020-115570GB-C22 and PID2020-115570GB-C21 funded by MCIN/AEI/10.13039/501100011033. Junta de Extremadura under project GR15068.

## References

1. David Camacho, Raúl Lara-Cabrera, J.J. Merelo-Guervós, Pedro A. Castillo, Carlos Cotta, Antonio J. Fernández-Leiva, Francisco Fernández de Vega, Francisco Chávez. "From ephemeral computing to deep bioinspired algorithms: New trends and applications", Future Generation Computer Systems, vol. 88, 2018, pp 735-746, ISSN 0167-739X, https://doi.org/10.1016/j.future.2018.07.056.
2. Díaz-Álvarez J, Castillo PÁ, Fernández de Vega F, Chávez F, Alvarado J. Population size influence on the energy consumption of genetic programming. Measurement and Control. 2022;55(1-2):102-115. doi:10.1177/00202940211064471
3. Eva García-Martín, Crefeda Faviola Rodrigues, Graham Riley, Håkan Grahn. "Estimation of energy consumption in machine learning" in Journal of Parallel and Distributed Computing, Volume 134, 2019, pp. 75-88, ISSN 0743-7315, https://doi.org/10.1016/j.jpdc.2019.07.007.
4. Fernández de Vega F., Díaz J., García J.Á., Chávez F., Alvarado J. "Looking for Energy Efficient Genetic Algorithms in Artificial Evolution". EA 2019. Lecture Notes in Computer Science, vol 12052, 2019, pp. 96-109, Springer, Cham. https://doi.org/10.1007/978-3-030-45715-0_8.

# Semantic Mutation Operator for Fast and Efficient Design of Bent Boolean Functions

Jakub Husa[0000−0003−0863−9952] and Lukáš Sekanina[0000−0002−2693−9011]

Faculty of Information Technology, Brno University of Technology,
Božetěchova 2/1, 612 00 Brno, Czech Republic
**ihusa@fit.vut.cz**

**Abstract.** Bent functions are a type of Boolean functions with properties that make them useful in cryptography. In this paper we propose a new semantic mutation operator for design of bent Boolean functions via genetic programming. To assess the efficiency of the proposed operator, we compare it to several other commonly used non-semantic mutation operators. Our results show that semantic mutation makes the evolutionary process more efficient, and significantly decreases the number of fitness function evaluations required to find a bent function.

**Keywords:** Genetic Programming · Semantic Mutation · Bent Boolean Functions.

## 1 Introduction

Boolean function is any function that takes some number of binary inputs and provides a single binary output. In symmetric cryptography Boolean functions often represent the only nonlinear element of a cipher [9].

To be usable for this purpose, the cryptographic function must fulfill several criteria, one of which is to have a high degree of nonlinearity (NL). The nonlinearity of a functions describes its Hamming distance to the nearest nearest affine function (which is a function that is either linear, or complementary to a linear function). To evaluate the nonlinearity of an $n$-input Boolean function $f$ its truth table needs to be converted to Walsh spectrum ($W_f$), using the Fast Walsh-Hadamard transform [7]. Walsh spectrum shows the correlation between the Boolean function and the set of all affine functions, and defines the nonlineaity of a Boolean function as follows:

$$NL(f) = 2^{n-1} - \frac{1}{2} \max_{\boldsymbol{x} \in \mathbb{F}_2^n}(W_f(\boldsymbol{x})) \tag{1}$$

The maximum nonlinearity of a function is limited by the number of its input variables as given by the following equation:

$$NL(f_{bent}) = 2^{n-1} - 2^{\frac{n}{2}-1} \tag{2}$$

This limit is known as the covering radius bound and can only be achieved by functions with an even number of variables. Functions that achieve this limit are known as bent Boolean functions. Notably, the nonlinearity of a function is invariant under all affine transformations, meaning that all of its variables are mutually interchangeable [1].

## 2   Methodology

To design the bent functions we use a heuristic method known as genetic programming (GP) whose syntactic tree is built out of nodes. Each of these nodes contains three genes, one that specifies the operation that is to be executed, and two that specify its inputs. Each input of a node can point to one of the Boolean function's variables, or to a new node. To avoid bloating of the tree, the maximum tree-depth is limited and the inputs of nodes in the bottom-most layer can only point to variables [9].

The population then undergoes an evolutionary process. Following $(1+\lambda)$ evolution strategy, in each generation the fittest and newest individual is selected to be a parent that generates $\lambda$ offspring through genetic mutation. Because the goal of this paper is to show the effects of genetic mutation, unlike most GP implementation, we do not use genetic crossover. Then, the fitness of the newly created individuals is evaluated, new best parent is selected, and this cycle repeats until the desired solution is found. In our case, the fitness of an individual is given by its nonlinearity (Eq. 1) and our goal is to create a function that is bent (Eq. 2).

When a gene is mutated, it is changed to a random valid value. This can mean a new randomly selected operation, a different input variable, or a newly generated node. Three commonly used (non-semantic) mutation operators are the uniform, point, and single mutation. Uniform mutation specifies a mutation rate, which is applied to every gene in the chromosome. Point mutation specifies a mutation count, and mutates the specified number of randomly selected genes. Single mutation selects a single active node and then mutates all three of its genes [9].

Our proposed semantic mutation operator is based on algebraic methods of bent function construction, and works as follows. First we randomly chose whether to mutate the structure or the function of the chromosome, and then randomly select one of two possible actions. If mutating the structure, the action can be to either activate or deactivate a new node. When activating a new node, a random input variable is selected and replaced with a new node. When deactivating a node, a random node is replaced with a variable. If mutating the function, the action can be to either shuffle all of the individual's variables or all of its operators.

When shuffling variables, the input variables of all nodes are replaced with new, randomly selected variables, with heavy bias towards selecting variables with fewest uses in order to create a more even distribution, as seen in Figure 1.

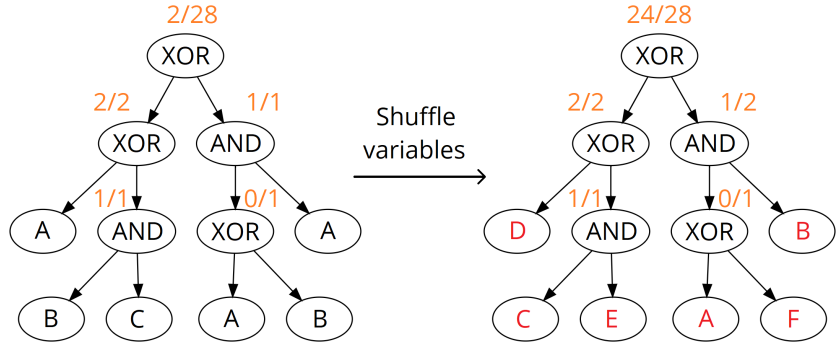

**Fig. 1.** Shuffle terminals - this action improves fitness by changing the distribution of terminal variables. Numbers above the nodes show their current/maximal nonlinearity.

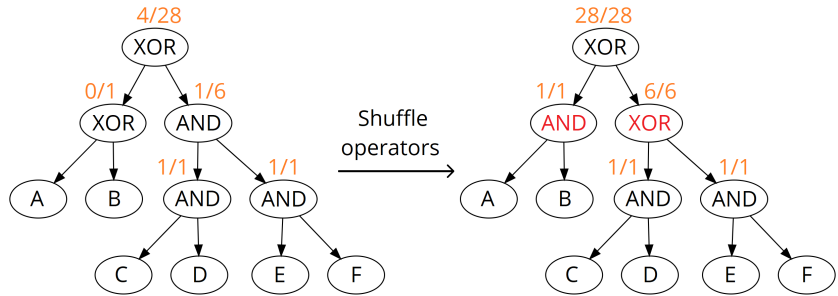

**Fig. 2.** Shuffle operators - this action improves fitness by changing operators of some of low-fitness nodes, while keeping the nodes with good fitness intact. Numbers above the nodes show their current/maximal nonlinearity.

When shuffling operators, the operations of all nodes are randomized with probability inversely proportional to their fitness (nonlinearity) to avoid disruptive mutations, as seen in Figure 2.

## 3   Results

In our experiments, we aim to design bent functions with $n = 12$ input variables, and measure the number of evaluations necessary to find the desired function. To make the comparison between the various mutation operators as fair as possible, all setting use the same set of operators {AND, XOR}, which is sufficient and highly effective for designing bent Boolean functions [2]. All settings also use the same tree depth $d = 7$, and population size $\lambda = 4$, which is unusually low for GP, but justified by the lack of genetic crossover, which makes larger populations unnecessary [3]. Based on our preliminary experiments, we select the best mutation rate $mr = 3\%$ used by uniform mutation, and $mc = 4$ used by point mutation.

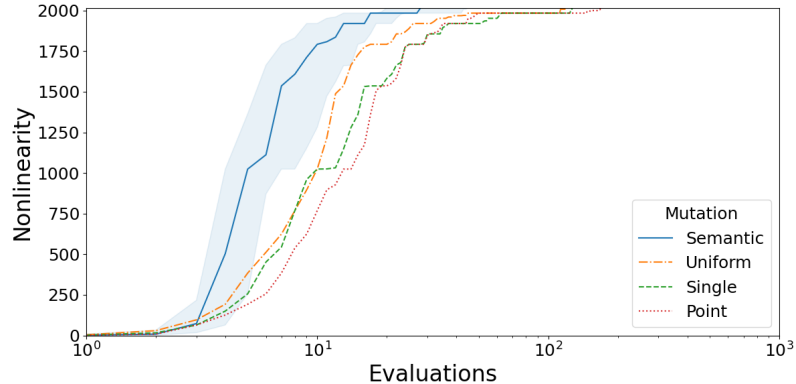

**Fig. 3.** Convergency curves for various types of mutation. The lines show the medians of 100 independent runs. The shaded area additionally shows the fist and third quartile of the semantic mutation. Note that the X-axis (evaluations) uses a logarithmic scale, and the Y-axis (fitness) terminates at the nonlinearity of an 12-input bent function ($NL(f_{bent}) = 2016$).

Using the four types of mutation, the process requires a median of 109 (semantic), 479 (uniform), 501 (single) or 673 (point) evaluations to find a bent function, as shown in Figure 3, which presents the convergency curves for the four examined types of mutation.

In conclusion, our results show that the proposed semantic mutation operator reduces the median number of fitness function evaluations necessary to find a bent function by 77.2% compared to the second-best operator, and thus significantly improves the efficiency of the evolutionary process, verified by non-parametric Mann-Whitney U-test at $\alpha = 0.05$.

**Acknowledgements:** This work was supported by the Czech Science Foundation Project 21-13001S and BUT Internal Grant Agency through project FITS-23-8141.

## References

1. Carlet, C.: Boolean Functions for Cryptography and Error-Correcting Codes, p. 257397. Encyclopedia of Mathematics and its Applications, Cambridge University Press (2010). https://doi.org/10.1017/CBO9780511780448.011
2. Husa, J., Dobai, R.: Designing bent boolean functions with parallelized linear genetic programming. In: Proceedings of the Genetic and Evolutionary Computation Conference Companion. pp. 1825–1832 (2017)
3. Husa, J., Sekanina, L.: Evolving cryptographic boolean functions with minimal multiplicative complexity. In: 2020 IEEE Congress on Evolutionary Computation (CEC). pp. 1–8. IEEE (2020)
4. ONeill, M.: Riccardo Poli, William B. Langdon, Nicholas F. McPhee: a field guide to genetic programming. Springer (2009)

# On the Performance of Evolutionary Algorithms with Unreliable Fitness Information⋆

Carlos Cotta[1,2][0000−0001−8478−7549]

[1] Dept. Lenguajes y Ciencias de la Computación, ETSI Informática,
Campus de Teatinos, Universidad de Málaga, 29071 Málaga, Spain
[2] ITIS Software, Universidad de Málaga, Spain
`ccottap@lcc.uma.es`

**Abstract.** We consider the use of evolutionary algorithms (EAs) in byzantine environments in which fitness information can be computed by malicious agents. The performance of panmictic EAs is analyzed in this context, measuring the influence of the rate of unreliability of the environment. It is shown that even for simple problems there is noticeable performance degradation, highlighting the need for appropriate mechanisms to cope with this issue.

**Keywords:** Evolutionary algorithms · Byzantine faults · Panmixia · Resilience

## 1 Introduction

Bioinspired optimization is a computationally intensive activity that is very much in need of abundant computational resources. In this sense, unconventional computational environments such as cloud, volunteer-computing and P2P networks have received great interest in recent years as adequate platforms to successfully solve complex problems [3]. The irregular and dynamic computational landscape that these platforms provide can pose a challenge to bioinspired algorithms though, thus underlining the need for algorithmic resilience [2]. Fortunately, bioinspired techniques in general and evolutionary algorithms (EAs) in particular are inherently quite resilient and are also flexible enough to being adapted for working in environments plagued with volatility or heterogeneity [6].

In this work we turn our attention to disruptions caused by malicious activities [8]. This kind of phenomena fall under the umbrella of byzantine failures, and can be described as cheating faults, whereby a contributor of computational resources purposefully alters the outcome of the computation by submitting erroneous results (aimed to feigning an activity or even to willingly damage the computation). These faults can have an impact on the algorithm depending on the components of the algorithm that are affected [4]. Empirically, it has been shown that EAs (particularly cellular EAs) can withstand certain types of byzantine failures [5]. We are here interested in analyzing both qualitatively and quantitatively what the effect of other types of byzantine failures can be. To this end, we initially focus on panmictic EAs.

⋆ This work is supported by Spanish Ministry of Science and Innovation under project Bio4Res (PID2021-125184NB-I00 – `http://bio4res.lcc.uma.es`) and by Universidad de Málaga, Campus de Excelencia Internacional Andalucía Tech

## 2 Algorithmic Setting

We consider an elitist generational EA with a panmictic population. This algorithm is used to optimize a certain objective function $f(\cdot)$. We will denote the values returned by this function as the *true fitness*. Let us further assume that a master-slave model [1] is used to compute fitness. Thus, we have a (possibly dynamic) network of computational nodes providing this fitness evaluation service. Some of these nodes are *cheaters* though. Similarly to [5], we consider a very simple model in which a fitness evaluation request returns an erroneous result with some probability $p$ (which will be later a control parameter in the experimentation). We will denote this result $\hat{f}^t(\cdot)$ as the *unreliable fitness*, where the superscript $t$ is used to indicate the time $t$ at which this value is returned (unreliable nodes are not assumed to consistently return the same erroneous result should the same solution been submitted for evaluation multiple times). Of course, there is no way of knowing a priori whether the value obtained when evaluating an individual is its true fitness or is a misleading value. Now, we consider two simple models of malicious behavior with regard to unreliable fitness:

- **randomizer**: the malicious computational node will return a value which is uncorrelated with the true fitness. This can be a random value within the range of the function, or -in our implementation- the true fitness value of another solution evaluated previously. This behavior would be related to nodes which want to feign an activity, without doing the actual work.
- **inverter**: in this case, the value returned is negatively correlated with the true fitness. In our experiments, we keep track of the maximum fitness $f_{\max}^t$ and minimum fitness $f_{\min}^t$ computed so far, and return the reflection of the true fitness within this interval, i.e., $\hat{f}^t(x) = f_{\max}^t - (f(x) - f_{\min}^t)$. This behavior would be related to nodes which want to actually inflict damage in the computation by leading the algorithm in the wrong direction.

## 3 Experimentation

We have conducted experiments with four objective functions, although due to space limitations we focus here on two of them, namely ONEMAX and LEADING-ONES. In both cases, we consider 100-bit solutions. As to the EA, it has a population of $\mu = 100$ individuals, uses binary tournament selection, single-point crossover ($p_X = .9$), bit-flip mutation ($p_M$ equivalent to a mutation rate $1/\ell$ per bit, where $\ell$ is the genome length), and elitist generational replacement. The total number of fitness evaluations is $10^6$. As to the unreliability rate $p$, since values $p > .5$ arguably correspond to pathological situations, we focus on

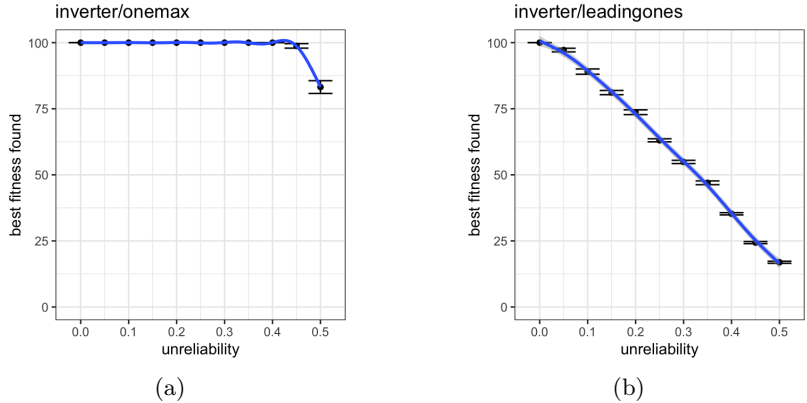

(a)                                            (b)

Fig. 1: Best true fitness of all solutions generated by the algorithm. (a) ONEMAX
(b) LEADING-ONES.

values between 0 and 0.5 in steps of 0.05, and use the results for $p = 0$ (reliable
scenario) to define the base-performance of the algorithm. We will focus on the
**inverter** model. We perform 50 runs for each parameter setting and problem.

Figs. 1(a)–(b) show the best true fitness found as a function of the unrelia-
bility probability $p$. Similarly to what was found in [5] for a scenario analogous
to **randomizer**, ONEMAX is barely affected by the unreliability factor $p$ except
in extreme cases. However, and quite distinctly, LEADING-ONES is much more
sensitive and not only fails to hit the optimum consistently for $p > 0$ but signifi-
cantly degrades as $p$ increases too. It must be noted that even though ONEMAX
seems less affected (ultimately due to its simplicity), the search dynamics of the
EA suffers greatly in the presence of unreliable fitness. If we define the relative
effort of the algorithm for a certain unreliability rate $p$ as the ratio between
the mean number of evaluations required to find a solution whose true fitness is
within some percentage of the optimum for that value of $p$ and for $p = 0$, we can
see in Fig. 2(a) how this effort noticeably grows up to nearly an order of magni-
tude for the larger values of $p$, and even precludes reaching fitness values close to
the optimal value. The presence of unreliable fitness information makes the EA
lose focus, as shown in Fig. 2(b) with a much larger population variance due to
the presence of *impostors* who survive by virtue of malicious fitness assignation.

## 4    Conclusions

We have here studied the impact that the presence of unreliable (and even ad-
versarial) fitness information has in the performance of panmictic EAs. It has
been shown that even for simple objective functions unreliability can pose se-
rious problems in terms of convergence and the effort required to attain good
quality solutions. We are currently working on mechanisms to cope with this issue. Ideas from noisy environments [7] can be serve as inspiration, although the nature of the phenomenon considered here is fundamentally different since uncertain fitness information does not gravitate around true fitness in this case.

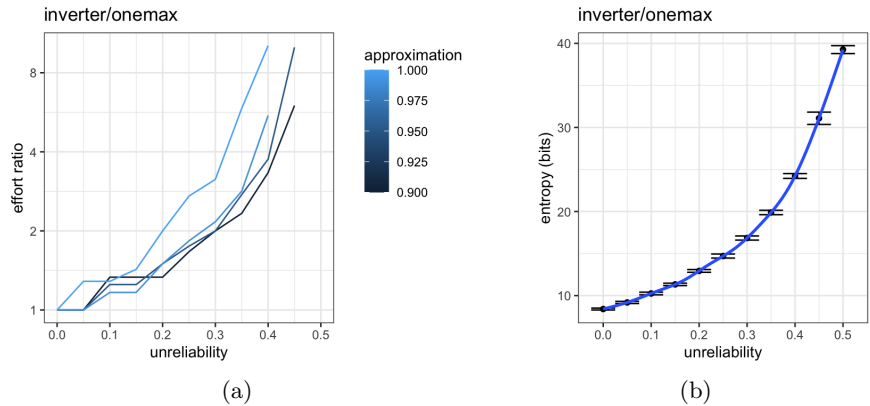

(a)                                   (b)

Fig. 2: (a) Relative effort of the algorithm to find a certain approximation of the optimum (lines are discontinued if the approximation cannot be found for a certain unreliability rate). Note the logarithmic y-axis. (b) Minimum population entropy during the run. In both cases, the results are for ONEMAX.

## References

1. Cantú-Paz, E.: Master-slave parallel genetic algorithms. In: Efficient and Accurate Parallel Genetic Algorithms, pp. 33–48. Springer US, Boston, MA (2001)
2. Cotta, C., Olague, G.: Resilient bioinspired algorithms: A computer system design perspective. In: Jiménez Laredo, J.L., et al. (eds.) Applications of Evolutionary Computation, Lecture Notes in Computer Science, vol. 13224, pp. 619–631. Springer, Cham (2022)
3. Mengistu, T.M., Che, D.: Survey and taxonomy of volunteer computing. ACM Computing Surveys **52**(3) (2019)
4. Muszyński, J., Varrette, S., Bouvry, P., Seredyński, F., Khan, S.U.: Convergence analysis of evolutionary algorithms in the presence of crash-faults and cheaters. Computers & Mathematics with Applications **64**(12), 3805–3819 (2012)
5. Muszyński, J., Varrette, S., Dorronsoro, B., Bouvry, P.: Distributed cellular evolutionary algorithms in a byzantine environment. In: 2015 IEEE International Parallel and Distributed Processing Symposium Workshop. pp. 307–313 (2015)
6. Nogueras, R., Cotta, C.: Analyzing self-⋆ island-based memetic algorithms in heterogeneous unstable environments. The International Journal of High Performance Computing Applications **32**(5), 676–692 (2018)
7. Rakshit, P., Konar, A., Das, S.: Noisy evolutionary optimization algorithms – A comprehensive survey. Swarm and Evolutionary Computation **33**, 18–45 (2017)
8. Sarmenta, L.F.: Sabotage-tolerance mechanisms for volunteer computing systems. Future Generation Computer Systems **18**(4), 561–572 (2002)

# Evolutionary Computation and Explainable AI: a year in review

Jaume Bacardit[1][0000−0002−2692−7205]
Alexander E.I. Brownlee[2][0000−0003−2892−5059]
Stefano Cagnoni[3][0000−0003−4669−512X]
Giovanni Iacca[4][0000−0001−9723−1830]
John McCall[5][0000−0003−1738−7056]
David Walker[6][0000−0001−8686−4253]

[1] Newcastle University, Newcastle u/o Tyne, UK `jaume.bacardit@newcastle.ac.uk`
[2] University of Stirling, Stirling, UK `alexander.brownlee@stir.ac.uk`
[3] University of Parma, Parma, Italy `stefano.cagnoni@unipr.it`
[4] University of Trento, Trento, Italy `giovanni.iacca@unitn.it`
[5] Robert Gordon University, Aberdeen, UK `j.mccall@rgu.ac.uk`
[6] University of Plymouth, Plymouth, UK `david.walker@plymouth.ac.uk`

**Abstract.** In 2022, we organized the first Workshop on Evolutionary Computation and Explainable AI (ECXAI). With no pretence at completeness, this paper briefly comments on its outcome, what has happened since then in the field, and our expectations for the near future.

**Keywords:** Explainable Artificial Intelligence · Evolutionary Computation and Optimisation · Machine Learning.

## 1 Explainable AI: facts and motivations

The increasing adoption of black-box algorithms, including Evolutionary Computation (EC)-based methods, has led to greater attention to generating explanations and their accessibility to end-users. This has created a fertile environment for the application of techniques in the EC domain for generating both end-user- and researcher-focused explanations. Furthermore, many XAI approaches in Machine Learning (ML) rely on search algorithms – e.g., [10] – that can draw on the expertise of EC researchers.

Important questions that automated decision-making techniques (such as EC and ML) have raised include: (1) Why has the algorithm obtained solutions in the way that it has? (2) Is the system biased? (3) Has the problem been formulated correctly? (4) Is the solution trustworthy and fair?

The goal of XAI and related research is to develop methods to interrogate AI processes and answer these questions. Our position is that, despite the differences in the problem formulation (ML vs. optimisation), using or adapting XAI techniques to explain EC-based processes that tackle search problems will improve such methods' accessibility to a wider audience, increasing their uptake and impact. As well as this, we posit that EC can play a crucial role in improving the state-of-the-art XAI techniques within the wider AI community.

Perhaps the most crucial reason why explainability is important is **trust**. The research community is already largely convinced of the value of EC approaches and is keen to increase the uptake of EC tools and methods by non-EC experts. Central to this is convincing users that they can trust the solutions that are generated by knowing *what* makes that solution better than something else, which might be synonymous with knowing *why* the solution was chosen.

Extending this theme is that of **validity**. EC methods, and optimisers in general, only optimise the target function. Explaining why a solution was chosen might clarify whether it is solving the actual problem or just exploiting an error or loophole in the problem's definition, which can lead to surprising or even amusing results [8], but can also simply yield frustratingly incorrect solutions.

EC is stochastic, which makes noise in the generated solutions virtually unavoidable. Thus, another motivation is whether we can explain which characteristics of the solution are crucial: its **malleability**. This property could be assessed by answering the question: "Which variables could be refined or amended for aesthetic or implementation purposes?".

Finally, when we define a problem, it is often hard to fully codify all the real-world goals of the system. We want something that is mathematically optimised but also something that corresponds to the problem owner's hard-to-codify intuition. By incorporating XAI into interactive EC, we could make it easier for the problem owner to interact with the optimiser [15].

Based mainly on these considerations and aiming to support these research lines, we decided to foster a tighter interaction between EC-based and other AI methods by organizing a dedicated workshop at GECCO 2022 [2], that will be iterated in 2023, inviting participants to focus their contributions on two main, complementary, issues, namely: (1) How EC can contribute to XAI?; and (2) How XAI can be used to explain EC-based solutions?

## 2   A year in review

Several papers were discussed at the ECXAI22 workshop. An introductory paper [2] outlined the broad research questions around EC and XAI, as noted above, and gave a brief literature review of recent work. Two papers explored routes for possible explainability in EC, one describing the mining of surrogate models for characteristics like the sensitivity of the objectives to each variable and inter-variable relationships for bitstring-encoded benchmark functions [13], and the other proposing Population Dynamics Plots to visualise the progress of an EA, to allow the lineage of solutions to be traced back to their origins and provide a route to explaining the behaviour of different algorithms [16].

Other papers explored the use of EC in providing explainability for ML systems, in particular, related to the assessment of Neural Transformers Trained on source code [11], the evolution of interpretable restriction policies for pandemics control [4], and the optimisation of explainable rule sets [12] and Learning Classifier Systems (LCS) through feature clustering [1].

### 2.1   Other recent relevant papers and new trends

Genetic Programming (GP) has long been claimed to produce better explainable models than other ML methods for its capacity to evolve complex symbolic expressions that intrinsically define their semantics. For the same reasons, GP has also the capacity to be a useful tool for post-hoc explanation of black-box models. These two viewpoints are extensively discussed in [9]. GP is also used in [14], aiming at deriving a context-aware approach, which essentially means developing a system that can decompose the main problem into a set of sub-problems (contexts) and finding specific solutions to each of them. According to the authors, this approach results in prediction models that are smaller and easier to interpret than those obtained by the evolutionary learning algorithms without context awareness.

Recent work has also focused on the use of evolutionary learning to induce decision trees combined with reinforcement learning (RL), both for discrete [5] and continuous action spaces [4]. A similar approach has been applied for the automated analysis of ultrasound imaging data [6], and, more recently, in the domain of multi-agent reinforcement learning (MARL) tasks [3].

One of the most appealing recent trends currently emerging relates to the use of Quality Diversity (QD) evolutionary algorithms, such as MAP-Elites, to search for a multitude of diverse policies for RL tasks [7].

## 3   An invitation and concluding remarks

The intersection between XAI and EC is evidently an emerging area, as demonstrated by the steady stream of recent publications and interest in the ECXAI workshop at GECCO 2022. Several, quite different approaches have been reported in the literature since then, and the topic is ripe for cross-fertilisation of ideas between the EC and XAI communities.

On the one hand, using XAI to attempt to explain the behavior and outcomes of EC techniques seems to be a viable way to attract attention on these optimisation methods and present them as a rigorous, robust alternative for solving complex optimisation problems. Using XAI in EC may also be a way to free the field from the excessive use of metaphors, that has been heavily criticized in the past few years, focusing more on the analysis of the algorithmic functionalities.

On the other hand, using EC for developing or augmenting XAI seems another important direction that deserves to be explored in the future. In this sense, the seminal works that have just been published on the use of QD algorithms for interpretable RL, or of EC for interpretable MARL, are encouraging.

We invite further participation in the ECXAI workshop at GECCO 2023[7].

## References

1. Andersen, H., Lensen, A., Browne, W.N.: Improving the search of learning classifier systems through interpretable feature clustering. In: Proceedings of the Genetic

---

and Evolutionary Computation Conference Companion. p. 1752–1756. GECCO '22, ACM, New York, NY, USA (2022)

2. Bacardit, J., Brownlee, A.E.I., Cagnoni, S., Iacca, G., McCall, J., Walker, D.: The intersection of evolutionary computation and explainable ai. In: Proceedings of the Genetic and Evolutionary Computation Conference Companion. p. 1757–1762. GECCO '22, Association for Computing Machinery, New York, NY, USA (2022)

3. Crespi, M., Custode, L.L., Iacca, G.: Towards interpretable policies in multi-agent reinforcement learning tasks. In: Bioinspired Optimization Methods and Their Applications: 10th International Conference, BIOMA 2022, Maribor, Slovenia, November 17–18, 2022, Proceedings. pp. 262–276. Springer, Cham (2022)

4. Custode, L.L., Iacca, G.: Interpretable AI for policy-making in pandemics. In: Proceedings of the Genetic and Evolutionary Computation Conference Companion. p. 1763–1769. GECCO '22, ACM, New York, NY, USA (2022)

5. Custode, L.L., Iacca, G.: Evolutionary learning of interpretable decision trees. IEEE Access **11**, 6169–6184 (2023)

6. Custode, L.L., Mento, F., Tursi, F., Smargiassi, A., Inchingolo, R., Perrone, T., Demi, L., Iacca, G.: Multi-objective automatic analysis of lung ultrasound data from COVID-19 patients by means of deep learning and decision trees. Applied Soft Computing **133**, 109926 (2023)

7. Ferigo, A., Custode, L.L., Iacca, G.: Quality diversity evolutionary learning of decision trees. arXiv:2208.12758 (2022)

8. Lehman, J., Clune, J., Misevic, D.: The surprising creativity of digital evolution: A collection of anecdotes from the evolutionary computation and artificial life research communities (2020), arXiv:1803.03453

9. Mei, Y., Chen, Q., Lensen, A., Xue, B., Zhang, M.: Explainable Artificial Intelligence by Genetic Programming: A survey. IEEE Transactions on Evolutionary Computation pp. 1–1 (2022), In press, Early Access

10. Ribeiro, M.T., Singh, S., Guestrin, C.: "Why should I trust you?". Explaining the predictions of any classifier. In: International Conference on Knowledge Discovery and Data Mining. ACM SIGKDD, New York, NY, USA (2016)

11. Saletta, M., Ferretti, C.: Towards the evolutionary assessment of neural transformers trained on source code. In: Proceedings of the Genetic and Evolutionary Computation Conference Companion. p. 1770–1778. GECCO '22, ACM, New York, NY, USA (2022)

12. Shahrzad, H., Hodjat, B., Miikkulainen, R.: Evolving explainable rule sets. In: Proceedings of the Genetic and Evolutionary Computation Conference Companion. p. 1779–1784. GECCO '22, ACM, New York, NY, USA (2022)

13. Singh, M., Brownlee, A.E.I., Cairns, D.: Towards explainable metaheuristic: Mining surrogate fitness models for importance of variables. In: Proceedings of the Genetic and Evolutionary Computation Conference Companion. p. 1785–1793. GECCO '22, ACM, New York, NY, USA (2022)

14. Tran, B., Sudusinghe, C., Nguyen, S., Alahakoon, D.: Building interpretable predictive models with context-aware evolutionary learning. Applied Soft Computing **132**, 109854 (2023)

15. Virgolin, M., De Lorenzo, A., Randone, F., Medvet, E., Wahde, M.: Model learning with personalized interpretability estimation (ML-PIE) (2021), arXiv:2104.06060

16. Walter, M.J., Walker, D.J., Craven, M.J.: An explainable visualisation of the evolutionary search process. In: Proceedings of the Genetic and Evolutionary Computation Conference Companion. p. 1794–1802. GECCO '22, ACM, New York, NY, USA (2022)